\documentclass[10pt,twocolumn,letterpaper]{article}
\usepackage{cvpr}
\usepackage[dvipsnames]{xcolor}
\usepackage{multirow}
\usepackage{acronym}
\usepackage{colortbl}
\usepackage{bm}
\usepackage{lipsum}
\usepackage{float}
\usepackage[misc]{ifsym}
\usepackage{dblfloatfix}

\usepackage{xspace}
\usepackage{setspace}
\usepackage[skip=3pt,font=small]{caption}
\makeatletter
\renewcommand{\paragraph}{%
  \@startsection{paragraph}{4}%
  {\z@}{1ex \@plus 1ex \@minus .2ex}{-1em}%
  {\normalfont\normalsize\bfseries}%
}
\makeatother

\definecolor{LightCyan}{rgb}{0.95,1,1}
\definecolor{Color}{rgb}{1,1,0.95}

\newcommand{\model}{MotionReFit\xspace}
\newcommand{\strategy}{MotionCutMix\xspace}
\newcommand{\dataset}[0]{\acs{dataset}\xspace}

\acrodef{llm}[LLM]{Large Language Model}
\acrodef{gan}[GAN]{Generative Adversarial Network}
\acrodef{mdm}[MDM]{Motion Diffusion Model}
\acrodef{slerp}[SLERP]{Spherical Linear Interpolation}
\acrodef{ddpm}[DDPM]{Denoising Diffusion Probabilistic Models}
\acrodef{mse}[MSE]{Mean-Squared Error}

\acrodef{dataset}[STANCE]{\underline{S}tyle \underline{T}ransfer, Fine-Grained \underline{A}djustme\underline{n}t, and Body Part Repla\underline{ce}ment}
\definecolor{cvprblue}{rgb}{0.21,0.49,0.74}
\usepackage[pagebackref,breaklinks,colorlinks,allcolors=cvprblue]{hyperref}

\title{Dynamic Motion Blending for Versatile Motion Editing\vspace{-6pt}}

\author{
    Nan Jiang\textsuperscript{1,2}\footnotemark[1] \quad
    Hongjie Li\textsuperscript{1}\footnotemark[1] \quad
    Ziye Yuan\textsuperscript{1}\footnotemark[1] \quad
    Zimo He\textsuperscript{1,2,3}\\
    Yixin Chen\textsuperscript{2} \quad 
    Tengyu Liu\textsuperscript{2} \quad
    Yixin Zhu\textsuperscript{1}$^{\,\textrm{\Letter}}$ \quad
    Siyuan Huang\textsuperscript{2}$^{\,\textrm{\Letter}}$
    \vspace{3pt}\\
    \small \textsuperscript{1} Institute for AI, Peking University \quad \textsuperscript{2} State Key Laboratory of General Artificial Intelligence, BIGAI \\
    \small \textsuperscript{3} Yuanpei College, Peking University \quad 
    \footnotemark[1]\;\;Equal contribution \quad 
    $\textrm{\Letter}$\,\,\texttt{yixin.zhu@pku.edu.cn,\,syhuang@bigai.ai}
    \vspace{3pt}\\
    \href{https://awfuact.github.io/motionrefit/}{https://awfuact.github.io/motionrefit/}
    \vspace{-21pt}
}

\begin{document}

\twocolumn[{%
    \renewcommand\twocolumn[1][]{#1}%
    \maketitle
    \begin{center}
        \centering
        \captionsetup{type=figure}
        \includegraphics[width=\linewidth,height=118pt]{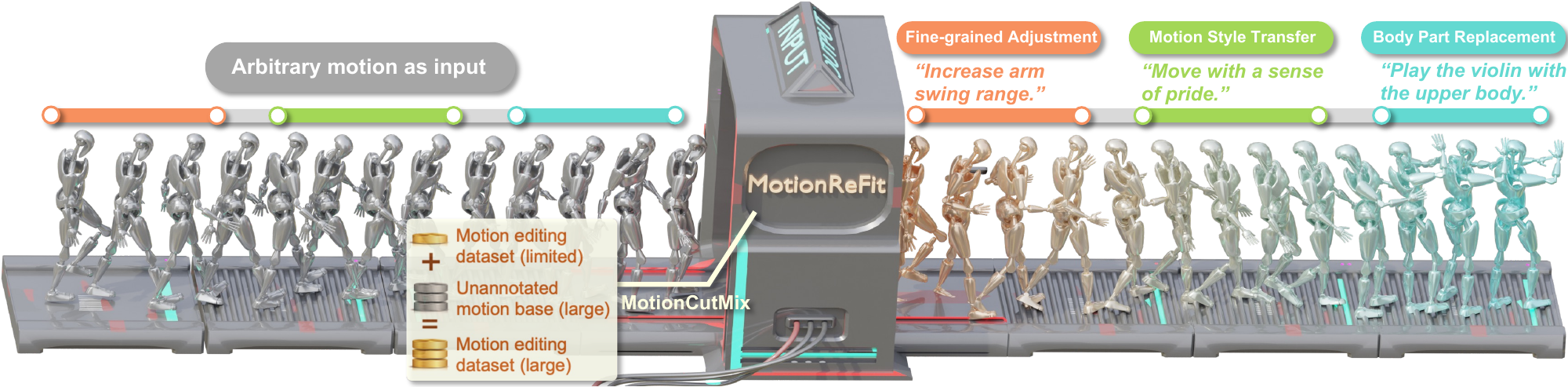}
        \captionof{figure}{\textbf{MotionReFit}, a universal framework for motion editing that handles various scenarios simply from textual guidance, offering both spatial and temporal editing capabilities. MotionReFit is supercharged with our proposed \textbf{MotionCutMix} training strategy, which leverages large-scale unannotated motion databases to augment the scarce motion editing triplets, enabling robust and generalizable editing.}
        \label{fig:teaser}
        \vspace{3pt}
    \end{center}%
}]

\begin{abstract}
Text-guided motion editing enables high-level semantic control and iterative modifications beyond traditional keyframe animation. Existing methods rely on limited pre-collected training triplets (original motion, edited motion, and instruction), which severely hinders their versatility in diverse editing scenarios. We introduce \strategy, an online data augmentation technique that dynamically generates training triplets by blending body part motions based on input text. While \strategy effectively expands the training distribution, the compositional nature introduces increased randomness and potential body part incoordination. To model such a rich distribution, we present \model, an auto-regressive diffusion model with a motion coordinator. The auto-regressive architecture facilitates learning by decomposing long sequences, while the motion coordinator mitigates the artifacts of motion composition. Our method handles both spatial and temporal motion edits directly from high-level human instructions, without relying on additional specifications or \acp{llm}. Through extensive experiments, we show that \model achieves state-of-the-art performance in text-guided motion editing. Ablation studies further verify that \strategy significantly improves the model's generalizability while maintaining training convergence. 
\end{abstract}

\section{Introduction}

Text-guided motion editing has emerged as a fundamental task in computer vision and animation~\cite{athanasiou2024motionfix,zhang2024finemogen,huang2024como}, enabling creators to perform \textit{semantic edits} (\eg, altering the right-hand movement to a circular motion) and \textit{style edits} (\eg, performing the motion in an angry style) through natural language instructions. Despite recent advances, current approaches~\cite{athanasiou2024motionfix,zhang2024finemogen,tevet2022mdm} face three critical limitations in achieving efficient, flexible, generalizable, and natural motion editing.

First, following InstructPix2Pix~\cite{brooks2023instructpix2pix}, existing methods~\cite{athanasiou2024motionfix,athanasiou2023sinc} rely on fixed triplets of original motion, edited motion, and editing instructions. This dependency severely restricts their ability to generalize across diverse scenarios, especially for style edits and novel motion-instruction combinations. Second, current models require explicit specification of body parts as auxiliary information, limiting their capability to autonomously comprehend high-level semantic instructions. Third, generating edited motions with smooth spatial-temporal transitions remains challenging.

To address these limitations, we introduce \textbf{\strategy}, a training technique that synthesizes novel triplets by blending body parts from multiple motion sequences. This approach leverages abundant unannotated motion data to augment expensive annotated editing triplets. Specifically, we employ a soft-mask mechanism for spatial blending of body parts, producing dynamically composited triplets of original motion, edited motion, and corresponding language instruction. This enables end-to-end editing using purely natural language input.

However, training with \strategy introduces two potential side-effects in motion generation: increased randomness and body part incoordination. To address these issues, we propose \textbf{\model} (\underline{Motion} \underline{RE}generation \underline{F}rom \underline{I}nput \underline{T}ext), an auto-regressive conditional diffusion model accompanied by a motion coordinator, as \cref{fig:teaser} shows. By employing an auto-regressive strategy, the motion is generated segment by segment, significantly facilitating convergence during training by decomposing long sequences. This approach also enables temporal editing with a smooth transition. To mitigate the incoordination in generated motion, we train a motion coordinator as a discriminator to assess whether a motion segment is the result of composition. This discriminator is used to refine the diffusion process as guidance, encouraging the generated motion segments to adherently resemble the pattern of original motions and avoiding model collapses to unnatural mode.

We extensively evaluate our approach using our proposed \textbf{\dataset} (\underline{S}tyle \underline{T}ransfer, Fine-Grained \underline{A}djustme\underline{n}t, and Body Part Repla\underline{ce}ment) dataset, which is developed for three text-guided motion editing tasks. Our experimental evaluations demonstrate that \model achieves high-fidelity edits across all three tasks while faithfully following the provided textual instructions. Through comprehensive ablation studies, we find that incorporating \strategy substantially enhances the model's generalization capability, particularly when training data is limited. Importantly, despite augmenting training data complexity, \strategy does not significantly impact the training convergence efficiency, allowing the model to benefit from expanded motion diversity without computational overhead.

Our primary contributions are threefold: 
\begin{itemize}
    \item We present \model, the first universal text-guided motion editing framework that achieves unrestricted editing capabilities for both body parts and temporal sequences. Powered by segmental motion synthesis mechanism and attention-based local-global refinement strategy, \model requires only original motion and editing instruction as input while delivering superior instruction adherence and motion naturalness.
    \item We introduce \strategy, a dynamic training technique that augments motion editing triplets online, enabling robust generalization, even with limited annotated data.
    \item We contribute \strategy, a motion-captured and manually annotated dataset for three editing tasks: body part replacement, fine-grained adjustment, and motion style transfer, providing diverse and high-quality examples for training and evaluation.
\end{itemize}

\begin{figure*}[ht!]
    \centering
    \begin{subfigure}{0.25\linewidth}
        \centering
        \includegraphics[width=\linewidth]{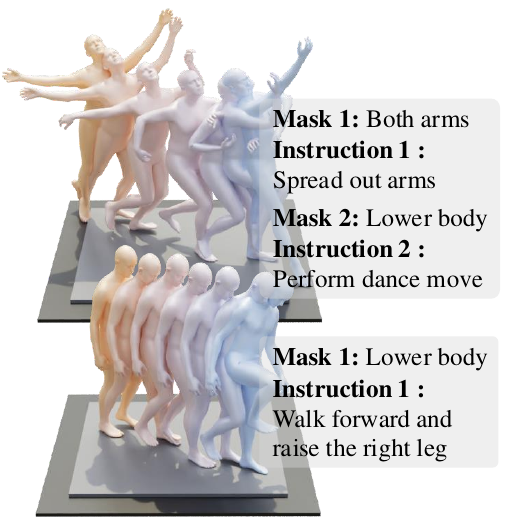}
        \caption{Body part replacement}
        \label{fig:datasets_a}
    \end{subfigure}
    \begin{subfigure}{0.442\linewidth}
        \centering
        \includegraphics[width=\linewidth]{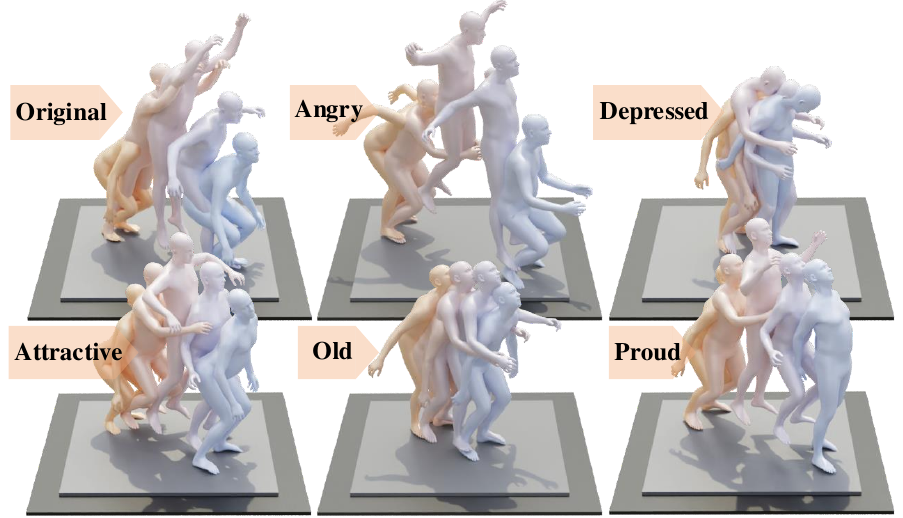}
        \caption{Motion style transfer}
        \label{fig:datasets_b}
    \end{subfigure}
    \begin{subfigure}{0.295\linewidth}
        \centering
        \includegraphics[width=\linewidth]{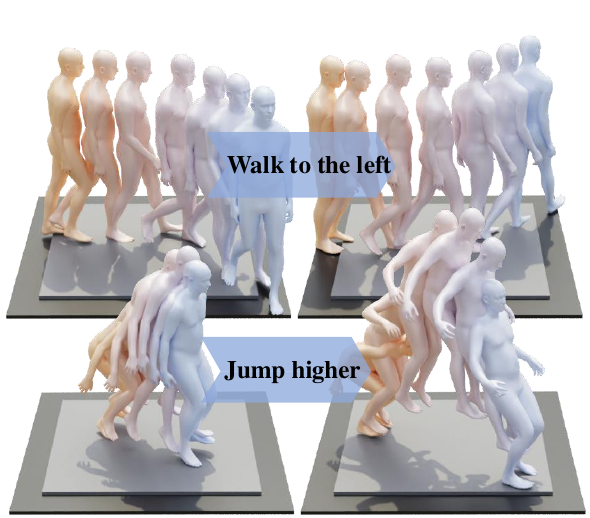}
        \caption{Fine-grained motion adjustment}
        \label{fig:datasets_c}
    \end{subfigure}
    \caption{\textbf{Sample sequences from our \dataset dataset.} Our work introduces three complementary datasets: (a) a body part replacement dataset comprising 13,000 sequences from HumanML3D~\cite{guo2022humanml3d}, annotated with an average of 2.1 body masks and corresponding motion descriptions; (b) a motion style transfer dataset containing 2 hours of new MoCap recordings that recreate HumanML3D sequences in various styles; and (c) a fine-grained motion adjustment dataset featuring 16,000 annotated triplets of generated motion pairs with their corresponding descriptions.}
    \label{fig:datasets}
\end{figure*}

\section{Related Work}

\paragraph{Data-Driven Motion Generation}

With access to large-scale motion datasets~\cite{shahroudy2016ntu,guo2022humanml3d,mahmood2019amass,punnakkal2021babel}, early motion generation approaches focused on predicting future motion~\cite{yan2018mt,aliakbarian2020stochastic}. Recent efforts have incorporated action labels and language descriptions to enhance the relevance and specificity of generated motions~\cite{tevet2022motionclip,hong2022avatarclip,petrovich2022temos,bie2022hit,harvey2020robust,wang2020learning,lin2018human,zhong2023attt2m}. The emergence of diffusion models~\cite{ho2020denoising,sohl2015deep} has marked a significant advancement in motion synthesis~\cite{tevet2022mdm,chen2023executing,dabral2023mofusion,zhang2024motiondiffuse,xu2023interdiff,dai2024motionlcm,zhang2024finemogen,jiang2023motiongpt,sun2024lgtm,jin2024act}. Several approaches~\cite{tevet2022mdm,zhang2024motiondiffuse,zhang2024finemogen,xie2023omnicontrol} have introduced motion editing capabilities. MDM~\cite{tevet2022mdm} supports part-level motion inpainting and temporal inbetweening, while FineMoGen~\cite{zhang2024finemogen} leverages \acp{llm} to interpret and execute editing instructions. However, these methods fail to simultaneously handle semantic and style edits.

\paragraph{Motion Style Transfer}

Early approaches in style transfer primarily relied on handcrafted features to address the complexities of defining and manipulating motion styles~\cite{unuma1995fourier, witkin1995motion, amaya1996emotion}. With the advent of deep learning, contemporary studies have favored data-driven techniques that leverage large datasets to extract and learn style features, utilizing approaches such as GAN~\cite{dong2020adult2child}, AdaIN~\cite{aberman2020unpaired}, and Diffusion~\cite{chang2022unifying,yin2024scalable,qian2024smcd}. While some methods employ neural networks trained on explicit pairs of original and edited motion styles~\cite{brand2000style,hsu2005style,wang2007multifactor,ikemoto2009generalizing,ma2010modeling,xia2015realtime} to directly translate specific movement patterns, others explore unpaired training strategies~\cite{aberman2020unpaired,tao2022style,huang2022unpaired,chang2022unifying,jang2022motion} to infer style from unaligned motion data or video inputs. However, despite these advancements in style transfer techniques, current methodologies predominantly address non-semantic motions and remain limited in their capacity to tailor arbitrary motions based on specific semantic textual descriptions.

\paragraph{Motion Editing} 

Motion editing, while sharing similarities with motion style transfer, remains comparatively under-explored. Early research focused on specific motion attributes such as adjusting motion paths~\cite{lockwood2011biomechanically,kim2009synchronized,gleicher2001motion}, adapting motions to different skeletal structures~\cite{aberman2020skeleton}, or altering motion-induced emotions~\cite{unuma1995fourier}.

In terms of semantic editing, \citet{tevet2022motionclip} and \citet{holden2016deep} proposed embedding motion sequences into latent vectors that encapsulate semantic information. However, this approach faces fundamental challenges as the embeddings may lack the fine-grained detail necessary for precise editing, and the latent space may not be sufficiently disentangled. Recent diffusion-based approaches~\cite{tevet2022mdm,zhang2024motiondiffuse,kim2023flame,pinyoanuntapong2024mmm} have enabled editing of existing motions through inpainting conditioned on textual instructions. However, these methods fix the joints of the remaining body parts, requiring clear delineation of the parts to be edited. 

Another significant line of research facilitates editing through motion composition, including temporal composition~\cite{athanasiou2022teach,shafir2024human,tseng2023edge,shi2024interactive}, spatial composition~\cite{athanasiou2023sinc,petrovich2022temos}, and comprehensive timeline control frameworks~\cite{petrovich2024multi}. Recent works such as FineMoGen~\cite{zhang2024finemogen}, Iterative Motion Editing~\cite{goel2024iterative}, and COMO~\cite{huang2024como} leverage foundation models for generating and editing motion, but they fail to handle arbitrary motion inputs without annotation. The work most similar to ours is TMED~\cite{athanasiou2024motionfix}, which employs a conditional diffusion model using both original motion and instructions as inputs, without requiring additional data. However, TMED's training on a limited set of triplets (original, edited, and instruction) hinders its generalizability to broader compositions, and it does not effectively handle temporal composition.

Addressing these limitations, our method provides an end-to-end solution that does not require additional user inputs while effectively handling a diverse range of motion-instruction compositions with the capability for both spatial and temporal edits.

\section{Problem Formulation and Representations}

\paragraph{Text-Guided Motion Editing}

Given an original motion sequence $\mathcal{M}_\text{ori}$ and an editing instruction $\mathcal{E}$ that specifies desired modifications, our goal is to generate an edited motion sequence $\mathcal{M}_\text{edit}$ that satisfies the following objectives:
\begin{itemize}
    \item $\mathcal{M}_\text{edit}$ should faithfully implement the modifications specified by $\mathcal{E}$, such as changes in motion style, intent, or specific body part movements.
    \item $\mathcal{M}_\text{edit}$ should maintain the integrity of $\mathcal{M}_\text{ori}$ by preserving aspects not explicitly specified by $\mathcal{E}$.
\end{itemize}

\paragraph{Human Motion Representations}

Our approach employs two complementary representations derived from the SMPL-X model~\cite{pavlakos2019expressive}. For direct motion manipulation, we use a keypoint-based representation $\mathcal{M^K} \in \mathbb{R}^{L \times N_K \times 3}$, where $L$ denotes sequence length and $N_K = 28$ represents the number of keypoints. These keypoints comprise 22 primary body joints from SMPL-X, supplemented by four finger joints (ring and index fingertips of both hands) for wrist pose determination, and two additional head joints to capture detailed head movements. In this representation, hands are treated as rigid bodies without detailed finger articulation. For compatibility with standard motion frameworks, we also utilize the SMPL-X parameter-based representation $\mathcal{M^S} = \{\mathbf{t}, \bm{\phi}, \mathbf{r} \}$. This representation consists of root translation $\mathbf{t} \in \mathbb{R}^{L \times 3}$, global orientation $\bm{\phi} \in \mathbb{R}^{L \times 3}$, and body pose parameters $\mathbf{r} \in \mathbb{R}^{L \times N_J \times 3}$, where $N_J = 21$ aligns with SMPL-X formulations. We use the mean body shape by setting $\bm{\beta}$ to zero.

These representations are interconvertible: Forward Kinematics transforms $\mathcal{M^S}$ to $\mathcal{M^K}$, while the reverse mapping uses a lightweight neural network followed by optimization to obtain $\mathcal{M^S}$ from $\mathcal{M^K}$. For simplicity, we omit representation superscripts when discussing motion in general terms. Details of motion representations and their conversions are in \cref{sec:keypoint-representation,sec:joint2smplx}, respectively.

\section{Training Data Construction}

This section details the construction of training triplets $\{\mathcal{M}_\text{ori},\mathcal{M}_\text{edit},\mathcal{E} \}$. We first present our proposed \dataset dataset in \cref{sec:proposed_dataset}. We then introduce a key motion composition operator in \cref{sec:spatial_motion_blending}, followed by the rules for constructing triplets across various editing settings in \cref{sec:dynamic_motion_composition}.

\subsection{STANCE Dataset}\label{sec:proposed_dataset}

Our \dataset dataset introduces three specialized components targeting common editing scenarios, as shown in \cref{fig:datasets}. Each component is carefully curated and verified by trained human annotators. Additional details for our \dataset dataset are available in \cref{sec:addtional-dataset-details}.

\paragraph{Body Part Replacement}

This editing type focuses on semantic edits where specific body parts are modified according to text instructions while preserving the motion of other parts. We improve upon previous approaches like~\cite{athanasiou2023sinc} that relied on \acp{llm} by having human annotators analyze rendered motions from the HumanML3D dataset~\cite{guo2022humanml3d} to assess body part participation. As illustrated in \cref{fig:datasets_a}, sequences can contain multiple mask sets, each annotated with descriptions of the masked body part's motion. We also introduce soft masks, detailed in \cref{sec:spatial_motion_blending}, to enable spatial blending.

\paragraph{Style Transfer}

As a type of style edit, this category aims to modify motion style without altering semantic content based on language instructions. We address the general case of style transfer across both locomotion and semantic motions. To overcome the lack of paired motions with identical semantics but different styles, we created a new MoCap dataset using the Vicon system. Professional actors recreated HumanML3D sequences in various styles (\eg, old, proud, depressed), resulting in 2 hours of motion comprising 750 stylized sequences.

\paragraph{Fine-grained Motion Adjustment}

This type of style edit enables detailed modifications without semantic changes (\eg, ``raise the right arm higher''). We introduce a novel approach that improves upon previous works like MotionFix~\cite{athanasiou2024motionfix}, which relied on TMR~\cite{petrovich2023tmr} representations for motion pairing. Instead, we utilize MLD~\cite{chen2023executing} as a text-to-motion generator to create 16 variants per instruction by perturbing the motion latent space. These variants are paired one-to-one, with human annotators describing the required transformations between pairs. After filtering out unnatural motions and unclear descriptions, we obtain 16,000 high-quality annotated triplets.

\subsection{Spatial Motion Blending}\label{sec:spatial_motion_blending}

\begin{figure}[t!]
    \centering
    \includegraphics[width=\linewidth]{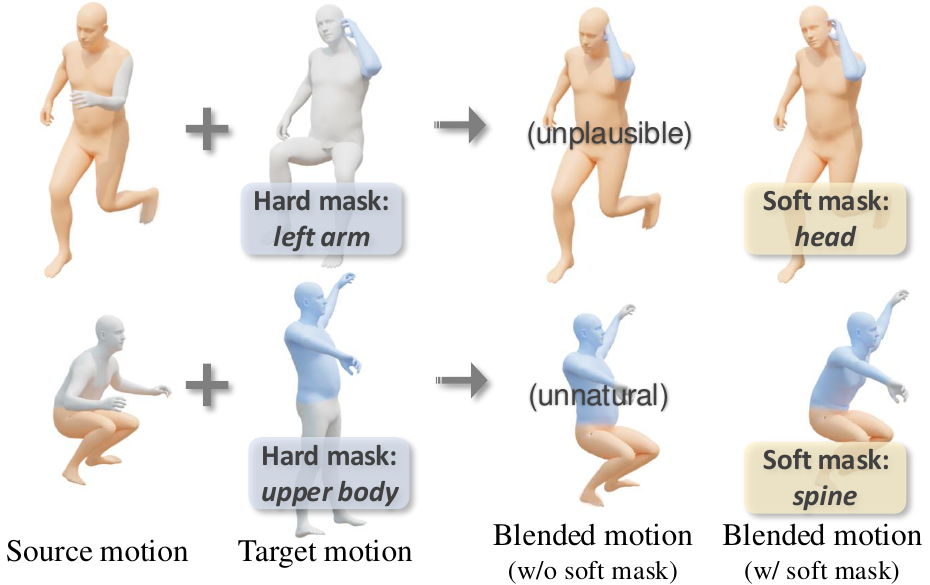}
    \caption{\textbf{Illustration of spatial motion blending.} We compare hard and soft masking approaches, showing how soft masks enable smoother transitions between body parts and eliminate unnatural artifacts at motion boundaries.}
    \label{fig:spatial_motion_blending}
\end{figure}

As illustrated in \cref{fig:spatial_motion_blending}, spatial motion blending enables the synthesis of novel motions by combining selected body parts from a source motion $\mathcal{M}_\text{src}$ with a target motion $\mathcal{M}_\text{tgt}$, guided by annotated masks. A mask is defined as $\mathbf{M}\subseteq\{0,1,\dots,N_j\}$, where $j\in\mathbf{M}$ indicates the $j^\text{th}$ joint (including pelvis) is selected. The blending process is guided by two annotated masks: a hard part $\mathbf{M}_\text{hard}$ and a soft part $\mathbf{M}_\text{soft}$, ensuring $\mathbf{M}_\text{hard}\cap\mathbf{M}_\text{soft}=\varnothing$. Joints within $\mathbf{M}_\text{hard}$ directly inherit rotations from $\mathcal{M}_\text{tgt}$, while those in $\mathbf{M}_\text{soft}$ undergo interpolation between source and target motions, ensuring smooth spatial transitions and motion coherence.

We denote the spatial motion blending process as $\text{BLD}(\mathcal{M}_\text{src},\mathcal{M}_\text{tgt},\{\mathbf{M}_\text{hard},\mathbf{M}_\text{soft}\})$. The resulting blended motion $\mathcal{M}_\text{bld} = \{ \mathbf{t}^\text{bld}, \bm{\phi}^\text{bld},\{ \mathbf{r}_j^\text{bld} \}_{j=1}^{N_J}\}$ is computed following these rules for each joint $j$:
\begin{equation*}
    \left\{
    \begin{array}{l@{\,}l@{\,}l}
        \mathbf{r}^\text{bld}_j = \mathbf{r}^\text{tgt}_j &\text{if} & j \in \mathbf{M}_\text{hard}\\
        \mathbf{r}^\text{bld}_j = \text{SLERP}(\mathbf{r}^\text{src}_j, \mathbf{r}^\text{tgt}_j, \alpha) &\text{if} & j \in \mathbf{M}_\text{soft}\\
        \mathbf{r}^\text{bld}_j = \mathbf{r}^\text{src}_j &\text{if} & j \notin \mathbf{M}_\text{hard}\,\text{and}\,j \notin \mathbf{M}_\text{soft}
    \end{array}
    \right.
\end{equation*}
where $\mathbf{r}^\text{src}$, $\mathbf{r}^\text{tgt}$, and $\mathbf{r}^\text{bld}$ represent joint rotations in the source, target, and blended motions respectively. The interpolation employs \ac{slerp} with a factor $\alpha$, which is randomly varied during training to increase motion diversity.

The global properties of the blended motion---orientation $\bm{\phi}^\text{bld}$ and translation $\mathbf{t}^\text{bld}$---are determined by the lower body motion. When the pelvis is included in $\mathbf{M}_\text{hard}$, the root pose follows $\mathcal{M}_\text{tgt}$; otherwise, it inherits from $\mathcal{M}_\text{src}$. This approach ensures consistency between the pelvis and the dominant lower body motion.

\subsection{MotionCutMix}\label{sec:dynamic_motion_composition}

We propose \strategy, a training technique that augments the limited motion data for training by leveraging variants from a larger motion database, which can be unannotated. Inspired by image augmentation~\cite{yun2019cutmix}, \strategy generates synthetic training samples through spatial motion blending on the training data. This enables the model to learn from diverse examples, capture high-level dependencies between original and edited motions, and enhance editing performance even with limited annotated training data.

\begin{figure*}
    \centering
    \includegraphics[width=\textwidth]{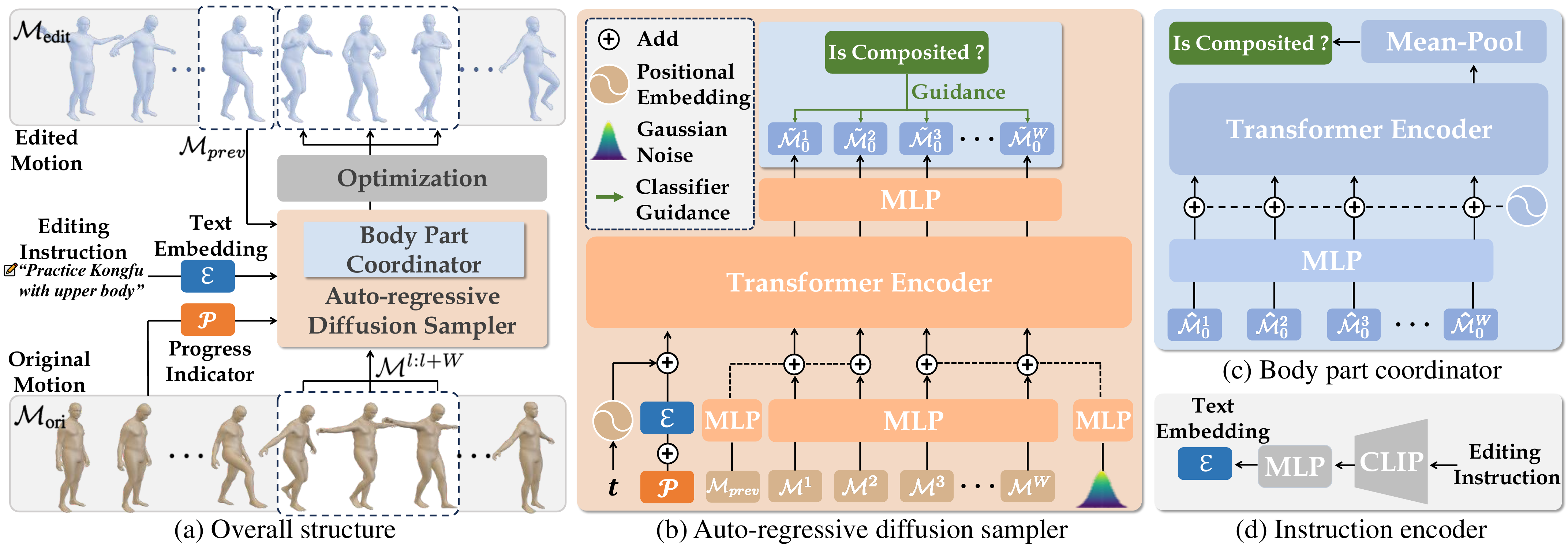}
    \caption{\textbf{Overview of \model.} Our auto-regressive approach processes the original motion through sliding windows, where body keypoints are encoded for input to a transformer-based motion diffusion model. To ensure motion continuity, noise is applied starting from the third frame while preserving the first two frames. The model incorporates an additional token integrating the editing instruction, diffusion step, and progress indicator. The generated keypoints undergo SMPL-X optimization and merging to create the final edited motion. To enhance body part coordination, we employ a discriminator trained to identify motion segments composed of multiple source motions, which guides the denoising process through classifier guidance.}
    \label{fig:method}
\end{figure*}

\strategy applies universally to both semantic and style edits, though with different composition rules. For semantic edits, \strategy randomly selects $\mathcal{M}_\text{src}$ from the large motion base and $\mathcal{M}_\text{tgt}$ from the dataset with body mask annotation. The training triplet $\{\mathcal{M}_\text{ori},\mathcal{M}_\text{edit},\mathcal{E} \}$ consists of $\mathcal{M}_\text{ori}=\mathcal{M}_\text{src}$ and $\mathcal{M}_\text{edit}=\text{BLD}(\mathcal{M}_\text{src},\mathcal{M}_\text{tgt},\mathbf{M}_\text{tgt})$, where $\mathbf{M}_\text{tgt}$ is the body mask annotated to $\mathcal{M}_\text{tgt}$. The editing instruction $\mathcal{E}$ is associated with $\mathbf{M}_\text{tgt}$, describing how the masked body part changes from $\mathcal{M}_\text{src}$ to $\mathcal{M}_\text{tgt}$.

Style edits present a different challenge since their parts requiring edits are already paired and cannot be randomly composited. To enable the model to learn generalized editing from limited data pairs, we split the editing into lower and upper bodies. For a source-target motion pair from the annotated dataset, \strategy randomly substitutes the non-edited body part of both $\mathcal{M}_\text{src}$ and $\mathcal{M}_\text{tgt}$ with the same motion sequence $\mathcal{M}_\text{ext}$ selected from an extra motion base. The blended pairs become $\mathcal{M}_\text{ori}=\text{BLD}(\mathcal{M}_\text{ext},\mathcal{M}_\text{src},\mathbf{M}_\text{edited-part})$ and $\mathcal{M}_\text{edit}=\text{BLD}(\mathcal{M}_\text{ext},\mathcal{M}_\text{tgt},\mathbf{M}_\text{edited-part})$, while $\mathcal{E}$ describes the style change on specific body parts.

\strategy effectively creates $N_L \times N_S$ original-edited pairs from $N_S$ annotated motion triplets, where $N_L$ denotes the size of the large motion base. By exposing the model to diverse motion combinations, \strategy enables better generalization and adherence to editing instructions.

\section{MotionReFit}

Our model performs end-to-end editing on arbitrary input motion by leveraging \strategy for creating training triplets. As shown in \cref{fig:method}, the framework consists of three key components: an auto-regressive motion diffusion model, a body part coordinator, and multiple condition encoders.

\subsection{Motion Diffusion Model}

At the core of our approach is an auto-regressive conditional diffusion model that generates edited motion segment by segment, guided by the original motion and text instruction. The model processes keypoint-based representations of human motion segments $\mathcal{M}^{l:l+W}$, where $l$ denotes the start frame and $W$ is the window size. For notation simplicity, we refer to $\mathcal{M}$ as ``the motion in the current segment'' throughout our discussion. Each segment $\mathcal{M}$ is transformed to a local coordinate system based on the root transformation of its initial frame, as detailed in \cref{sec:keypoint-canonicalization}.

Following the \ac{ddpm}~\cite{ho2020denoising} framework, we implement a forward diffusion process as a Markov Chain that progressively adds noise to clean edited motion segments $\mathcal{M}_\text{edit}$ over $T$ steps. Using $\mathcal{M}_t$ to denote the noisy version of $\mathcal{M}_\text{edit}$ at diffusion step $t$, the noise addition process follows:
\begin{equation}
    q(\mathcal{M}_{t} | \mathcal{M}_{t-1}) = \mathcal{N}(\mathcal{M}_{t}; \sqrt{1 - \beta_t} \mathcal{M}_{t-1}, \beta_t \mathbf{I}),
\end{equation}
where $\beta_t \in (0, 1)$ is a variance schedule controlling noise magnitude per step, and $\mathbf{I}$ is the identity matrix.

The reverse denoising process is learned by a network $\epsilon_\theta$ (\cref{sec:module-details}), which sequentially denoises samples across $T$ steps starting from $\mathcal{M}_{T} \sim \mathcal{N}(\mathbf{0}, \mathbf{I})$. Following \citet{ho2020denoising}, we train the model by minimizing the \ac{mse} between predicted and added noise:
\begin{equation}
    \mathcal{L}=\mathbb{E}_{\mathcal{M}_\text{0}\sim q(\mathcal{M}_\text{0}|\mathcal{C}),t\sim[1,T]}||\epsilon - \epsilon_\theta(\mathcal{M}_t, t, \mathcal{C})||_{2}^2.
\end{equation}

The conditional terms $\mathcal{C} = \{\mathcal{M}_\text{prev},\mathcal{M}_\text{ori},\mathcal{E}, \mathcal{P}\}$ comprise:
(i) two frames of motion $\mathcal{M}_\text{prev}$ right before the current segment, encoded via MLP without noise processes;
(ii) the original motion segment $\mathcal{M}_\text{ori}$;
(iii) the editing instruction encoded through CLIP~\cite{radford2021learning}; and
(iv) a progress indicator $\mathcal{P}$ representing the normalized starting frame position within the edited motion~\cite{jiang2024scaling} using sinusoidal positional encoding~\cite{vaswani2017attention}.

\begin{figure*}[b!]
    \centering
    \includegraphics[width=\textwidth]{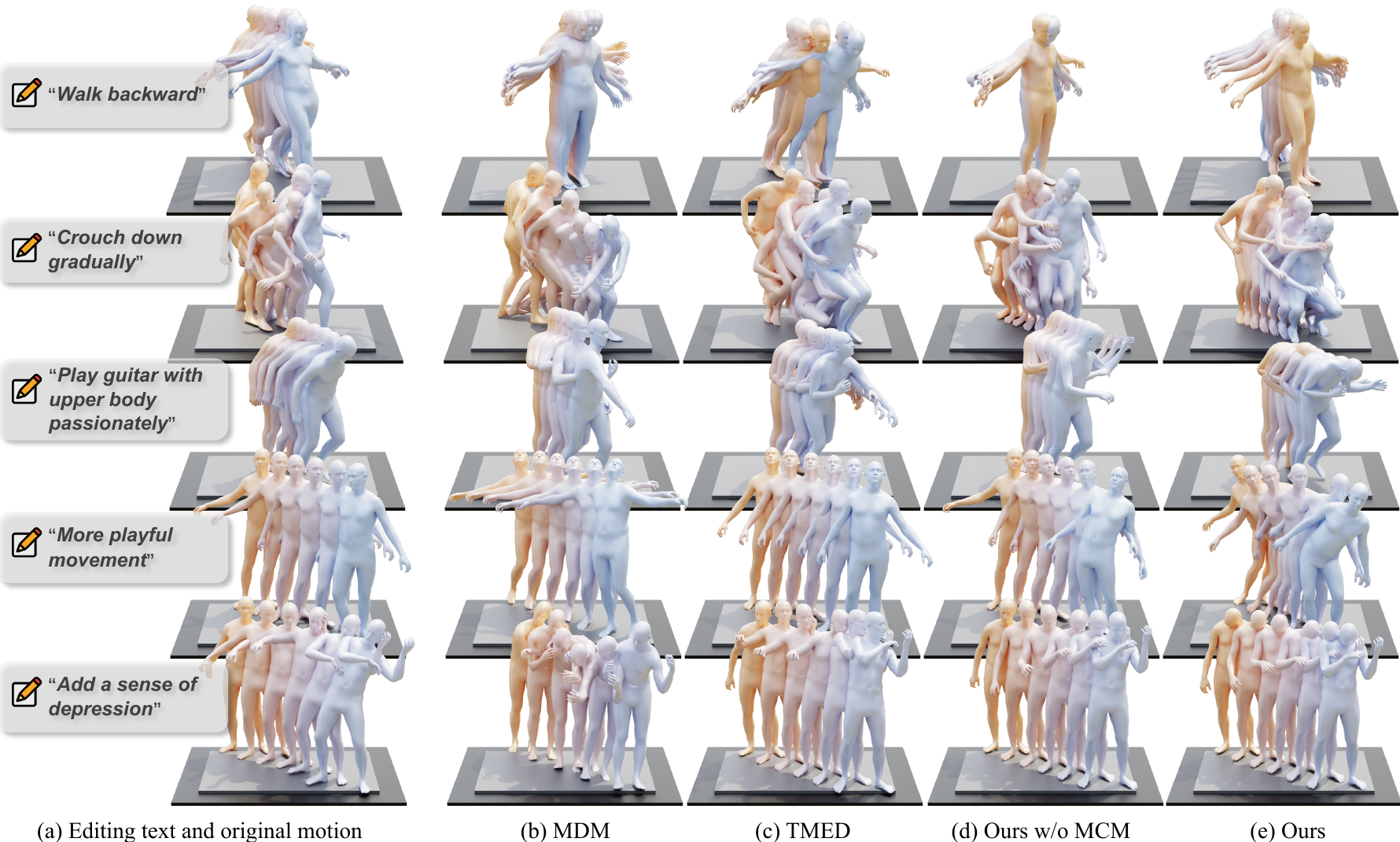}
    \caption{\textbf{Qualitative comparison of text-guided motion editing results.} Each sequence shows the original motion alongside edits by \model and baseline methods. Motion trajectories are visualized with a color gradient from \textcolor{BurntOrange}{\textbf{orange}} (starting position) to \textcolor{Periwinkle}{\textbf{blue}} (ending position), with spatial offsets applied to emphasize motion differences.}
    \label{fig:compare}
\end{figure*}

To strengthen the model's adherence to editing instructions, we use classifier-free guidance~\cite{ho2022classifier} with weight $w$:
\begin{equation}
    \tilde{\epsilon}_{\theta}(\mathcal{M}_t, t, \mathcal{C})=(1+w)\epsilon_{\theta}(\mathcal{M}_t, t, \mathcal{C})-w\epsilon_{\theta}(\mathcal{M}_t, t, \mathcal{C}'),
\end{equation}
where $\mathcal{C}'=\{\mathcal{M}_\text{prev},\mathcal{M}_\text{ori},\varnothing ,\mathcal{P}\}$ represents the conditional terms with the instruction removed.

\subsection{Body Part Coordinator}\label{sec:body_part_coordinating}

Training on composed motion data introduces a critical challenge: generated motions may exhibit incorrect coordination patterns, such as synchronized movement of same-side feet and hands during walking. To address this, we introduce a motion discriminator $D$ that provides classifier guidance to the diffusion model, ensuring natural coordination between body parts.

The discriminator is trained to classify motion segments as either coherent (uncomposited) or artificially composed. We construct a training dataset where 50\% of samples come from unmodified source motion segments in the HumanML3D dataset~\cite{guo2022humanml3d}, while the remaining 50\% are synthetically created by compositing body parts from different motion segments. Through this balanced training approach, the discriminator learns to identify subtle coordination patterns that distinguish natural from composited motions.

\begin{table*}[t!]
    \centering
    \setlength{\tabcolsep}{3pt}
    \caption{\textbf{Quantitative comparison across body part replacement (upper) and style transfer (lower) tasks.} Each metric reports mean over 10 evaluations with 95\% confidence intervals ($\pm$). Arrows ($\rightarrow$) indicate metrics where values closer to real data are better. \textbf{Bold} denotes best performance.}
    \label{tab:results-main}
    \resizebox{\linewidth}{!}{%
        \begin{tabular}{lccccccccccc}
            \toprule
            \multirow{2}{*}{Method} & \multirow{2}{*}{FID$\downarrow$} & \multirow{2}{*}{Diversity$\rightarrow$} & \multirow{2}{*}{FS$\downarrow$} & \multicolumn{4}{c}{Edited-to-Source Retrieval} & \multicolumn{4}{c}{Edited-to-Target Retrieval} \\
            \cmidrule(r){5-8} \cmidrule(l){9-12} & & & & R@1$\rightarrow$ & R@2$\rightarrow$ & R@3$\rightarrow$ & AvgR$\rightarrow$ & R@1$\uparrow$ & R@2$\uparrow$ & R@3$\uparrow$ & AvgR$\downarrow$ \\
            \midrule
            Real Data & $0.01^{\pm .001}$& $36.06^{\pm .436}$ & $0.98^{\pm .000}$ & $52.08^{\pm .371}$ & $54.32^{\pm .314}$ & $56.00^{\pm .365}$ &  $8.28^{\pm .045}$ & $100.0^{\pm .000}$ & $100.0^{\pm .000}$ & $100.0^{\pm .000}$ & $1.00^{\pm .000}$ \\
            MDM-BP~\cite{tevet2022mdm} & $0.44^{\pm .030}$ & $36.71^{\pm .701}$ & $0.91^{\pm .003}$ & $69.11^{\pm .912}$ & $79.75^{\pm .711}$ & $85.14^{\pm .561}$ & $2.20^{\pm .028}$ & $39.05^{\pm .469}$ & $46.39^{\pm .441}$ & $50.57^{\pm .556}$ & $8.92^{\pm .033}$ \\
            TMED~\cite{athanasiou2024motionfix} & $0.52^{\pm .034}$ & $35.37^{\pm .540}$ & $0.90^{\pm .008}$ & $38.59^{\pm 1.169}$ & $44.10^{\pm .932}$ & $48.67^{\pm .911}$ & $9.31^{\pm .211}$ & $42.70^{\pm 1.533}$ & $52.89^{\pm 1.286}$ & $58.32^{\pm 1.430}$ & $6.47^{\pm .118}$ \\
            TMED w/ MCM & $0.54^{\pm .028}$ & $35.67^{\pm .482}$ & $0.90^{\pm .006}$ & $41.29^{\pm .631}$ & $46.13^{\pm .881}$ & $49.80^{\pm .945}$ & $9.38^{\pm .095}$ & $50.62^{\pm 1.612}$ & $61.95^{\pm 1.421}$ & $68.52^{\pm 1.484}$ & $\pmb{4.48^{\pm .119}}$ \\
            Ours w/o MCM & $0.23^{\pm .026}$ & $36.34^{\pm .620}$ & $0.96^{\pm .003}$ & $93.17^{\pm .273}$ & $96.30^{\pm .178}$ & $97.33^{\pm .206}$ & $1.27^{\pm .011}$ & $51.18^{\pm .206}$ & $53.71^{\pm .275}$ & $55.30^{\pm .371}$ & $8.51^{\pm .020}$ \\
            Ours w/o BC & $0.23^{\pm .016}$ & $36.18^{\pm .523}$ & $\pmb{0.97^{\pm .003}}$ & $52.51^{\pm .595}$ & $\pmb{56.03^{\pm .368}}$ & $58.19^{\pm .358}$ & $\pmb{7.54^{\pm .038}}$ & $60.78^{\pm .471}$ & $67.17^{\pm .457}$ & $71.11^{\pm .521}$ & $4.74^{\pm .042}$ \\
            Ours full & $\pmb{0.20^{\pm .025}}$ & $\pmb{36.01^{\pm .758}}$ & $\pmb{0.97^{\pm .002}}$ & $\pmb{52.48^{\pm .337}}$ & $56.13^{\pm .361}$ & $\pmb{58.59^{\pm .234}}$ & $7.46^{\pm .034}$ & $\pmb{61.37^{\pm .457}}$ & $\pmb{68.35^{\pm .493}}$ & $\pmb{72.20^{\pm .314}}$ & $4.65^{\pm .029}$ \\
            \midrule
            Real Data & $0.01^{\pm .001}$ & $33.98^{\pm .865}$ & $0.98^{\pm .000}$ & $50.94^{\pm 1.791}$ & $62.88^{\pm .925}$ & $67.40^{\pm .828}$ & $6.28^{\pm .058}$ & $100.0^{\pm .000}$ & $100.0^{\pm .000}$ & $100.00^{\pm .000}$ & $1.00^{\pm .000}$ \\
            MDM-BP~\cite{tevet2022mdm} & $0.39^{\pm .033}$ & $\pmb{33.64^{\pm .835}}$ & $0.89^{\pm .010}$ & $62.40^{\pm 1.977}$ & $82.78^{\pm 1.100}$ & $89.62^{\pm 1.156}$ & $1.96^{\pm .062}$ & $38.89^{\pm 2.152}$ & $53.51^{\pm 1.167}$ & $60.24^{\pm 1.122}$ & $7.14^{\pm .071}$ \\
            TMED~\cite{athanasiou2024motionfix} & $1.54^{\pm .093}$ & $34.37^{\pm 1.111}$ & $0.90^{\pm .010}$ & $28.44^{\pm 1.156}$ & $40.03^{\pm 1.173}$ & $46.53^{\pm 1.280}$ & $8.48^{\pm .104}$ & $24.76^{\pm 1.440}$ & $38.33^{\pm 2.067}$ & $45.62^{\pm .934}$ & $8.12^{\pm .099}$ \\
            TMED w/ MCM & $0.84^{\pm .060}$ & $34.35^{\pm .669}$ & $0.92^{\pm .004}$ & $39.83^{\pm 1.522}$ & $55.00^{\pm 1.608}$ & $\pmb{62.92^{\pm 1.463}}$ & $5.37^{\pm .112}$ & $33.02^{\pm 1.024}$ & $47.60^{\pm 1.303}$ & $56.94^{\pm 1.242}$ & $6.15^{\pm .072}$ \\
            Ours w/o MCM & $0.23^{\pm .017}$ & $34.05^{\pm 1.077}$ & $0.93^{\pm .006}$ & $87.05^{\pm 1.345}$ & $98.33^{\pm .556}$ & $99.41^{\pm .313}$ & $1.16^{\pm .012}$ & $51.39^{\pm 1.406}$ & $63.58^{\pm 1.058}$ & $67.88^{\pm .699}$ & $7.15^{\pm .102}$ \\
            Ours w/o BC & $0.16^{\pm .018}$ & $34.51^{\pm .681}$ & $\pmb{0.95^{\pm .003}}$ & $45.52^{\pm 1.146}$ & $57.05^{\pm 1.120}$ & $62.29^{\pm .810}$ & $6.57^{\pm .080}$ & $62.26^{\pm 1.838}$ & $74.69^{\pm .814}$ & $79.90^{\pm 1.227}$ & $3.51^{\pm .081}$ \\
            Ours full & $\pmb{0.14^{\pm .015}}$ & $34.19^{\pm .865}$ & $0.94^{\pm .004}$ & $\pmb{47.67^{\pm 1.099}}$ & $\pmb{57.71^{\pm 1.039}}$ & $62.50^{\pm .439}$ & $\pmb{6.46^{\pm .086}}$ & $\pmb{63.82^{\pm 1.551}}$ & $\pmb{76.35^{\pm .988}}$ & $\pmb{80.69^{\pm 1.009}}$ & $\pmb{3.48^{\pm .062}}$ \\
            \bottomrule
        \end{tabular}
    }%
\end{table*}

During the motion generation process, we integrate the trained discriminator as a classifier guidance:
\begin{equation}
    \tilde{\mathcal{M}}_\text{0} = \hat{\mathcal{M}}_{0} + \lambda \nabla_{\hat{\mathcal{M}}_\text{0}} D(\hat{\mathcal{M}}_\text{0}),
\end{equation}
where $\hat{\mathcal{M}}_\text{0}=\tilde{\epsilon}_{\theta}(\mathcal{M}_t, t, \mathcal{C})$ is the model's output, $\tilde{\mathcal{M}}_\text{0}$ represents the motion segment after applying classifier guidance, $\lambda$ controls the guidance strength, and $\nabla_{\hat{\mathcal{M}}_\text{0}} D(\hat{\mathcal{M}}_\text{0})$ is the discriminator's gradient with respect to $\hat{\mathcal{M}}_\text{0}$. To refine body part coordination while preserving the overall motion structure, we apply this classifier guidance during the final 20 steps of the auto-regressive sampling process.

\section{Experiments}

\subsection{Evaluation Settings}

\paragraph{Tasks and Datasets}

Our main experiments evaluate two key tasks: body part replacement (semantic edits) and style transfer (style edits), as detailed in \cref{sec:proposed_dataset}. We assess all methods using our task-specific datasets, split into training (80\%), validation (5\%), and testing (15\%) sets. For training data preparation, we create triplets (original motion, edited motion, instruction) from our \dataset dataset using composition rules in \cref{sec:dynamic_motion_composition}. The training set of HumanML3D~\cite{guo2022humanml3d} serves as our extensive motion base for \strategy implementations. The evaluation of fine-grained adjustment capabilities is presented separately in \cref{sec:results-of-adjustment}.

Additionally, we evaluate our method on the MotionFix dataset~\cite{athanasiou2024motionfix}. For these experiments, we disable \strategy and configure our auto-regressive diffusion model with a 16-frame window size. 

\paragraph{Baeslines} 

We compare our method against two text-guided motion editing baselines: MDM-BP~\cite{tevet2022mdm} and TMED~\cite{athanasiou2024motionfix}. MDM-BP extends the original MDM by incorporating body-part inpainting and ground-truth body part information to specify fixed and edited parts. For TMED comparisons, we maintain their original experimental settings (detailed in \cref{sec:baseline-adaption}).

\paragraph{Ablations} 

We conduct the following ablation studies to analyze key components of our method:
\begin{itemize}
    \item Ours w/o MCM: To isolate the impact of motion composition, we evaluate our method using fixed original-edited pairs following SINC~\cite{athanasiou2023sinc}, without \strategy during training.
    \item  TMED~\cite{athanasiou2024motionfix} w/ MCM: To assess \strategy's broader applicability, we integrate it into TMED's~\cite{athanasiou2024motionfix} training pipeline.
    \item Ours w/o BC: To validate our body part coordinator, we evaluate our method without the body part coordinator from \cref{sec:body_part_coordinating}.
    \item MotionCutMix Ratio: To examine data composition effects, we vary the proportions of motion base data used in \strategy.
    \item Annotated Data Size: To evaluate \strategy with limited annotations, we train models with different proportions of annotated data.
    \item Window Size: To optimize temporal processing, we experiment with different sliding window sizes for auto-regressive generation.
    \item Training steps: To assess data randomness effects on convergence, we track performance under different \strategy ratios during training.
\end{itemize}

\paragraph{Metrics}

We employ the Edited-to-Source Retrieval (E2S) and Edited-to-Target Retrieval (E2T) scores from \citet{athanasiou2024motionfix}, using TMR~\cite{petrovich2023tmr} features. We report R@1, R@2, R@3, and AvgR with 32-batch random gallery sampling from the test set. For quality and diversity assessment, we use Fréchet Inception Distance (FID), Foot Score (FS)~\cite{zhao2023synthesizing}, Diversity, and Multimodality~\cite{tevet2022mdm}. E2S interpretation varies by task: high scores are desired for fine-grained adjustments and MotionFix evaluations, while body part replacement and style transfer should match reference dataset distributions (detailed in \cref{sec:e2s-score}).

\begin{figure*}[ht!]
    \centering
    \begin{subfigure}{0.25\linewidth}
        \centering
        \includegraphics[width=\linewidth]{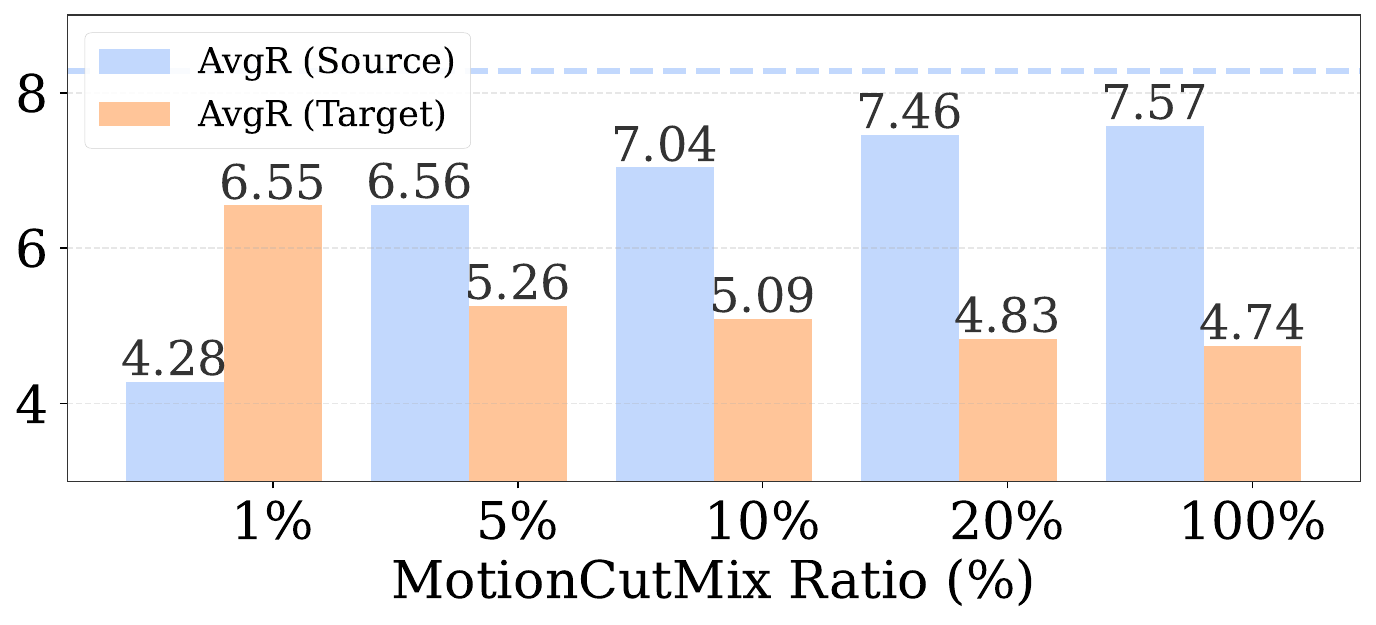}
        \caption{}
        \label{fig:ablation_a}
    \end{subfigure}%
    \begin{subfigure}{0.25\linewidth}
        \centering
        \includegraphics[width=\linewidth]{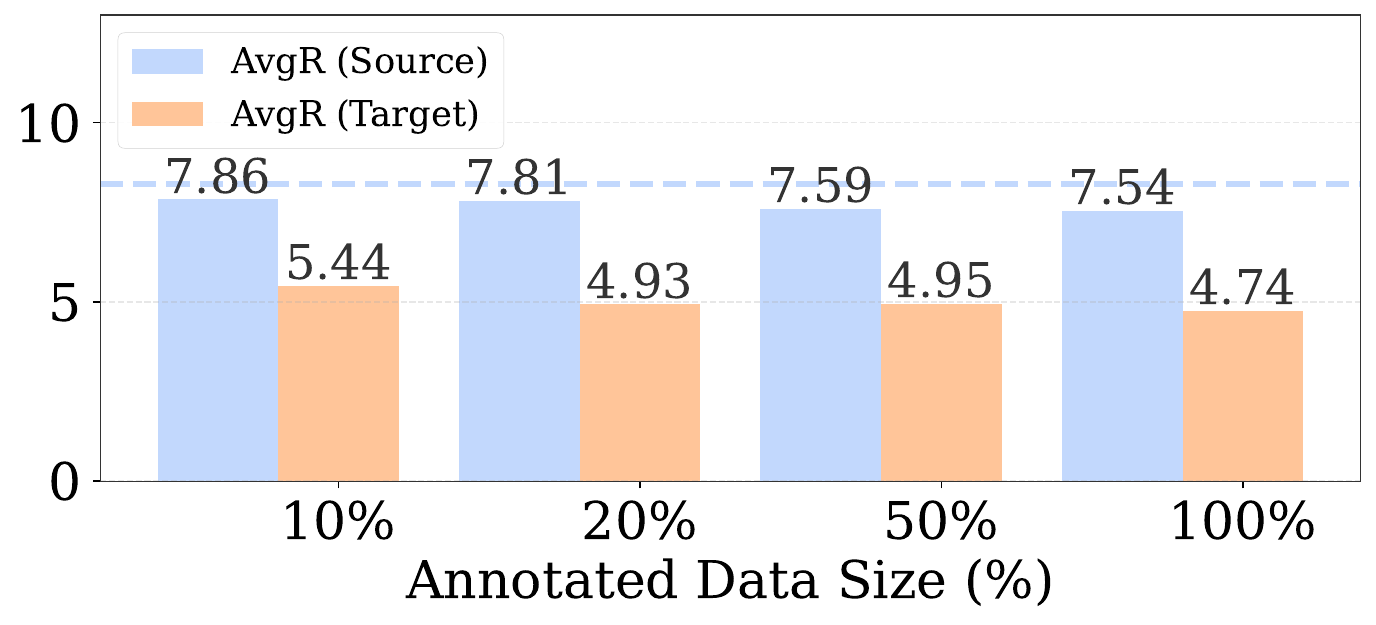}
        \caption{}
        \label{fig:ablation_b}
    \end{subfigure}%
    \begin{subfigure}{0.25\linewidth}
        \centering
        \includegraphics[width=\linewidth]{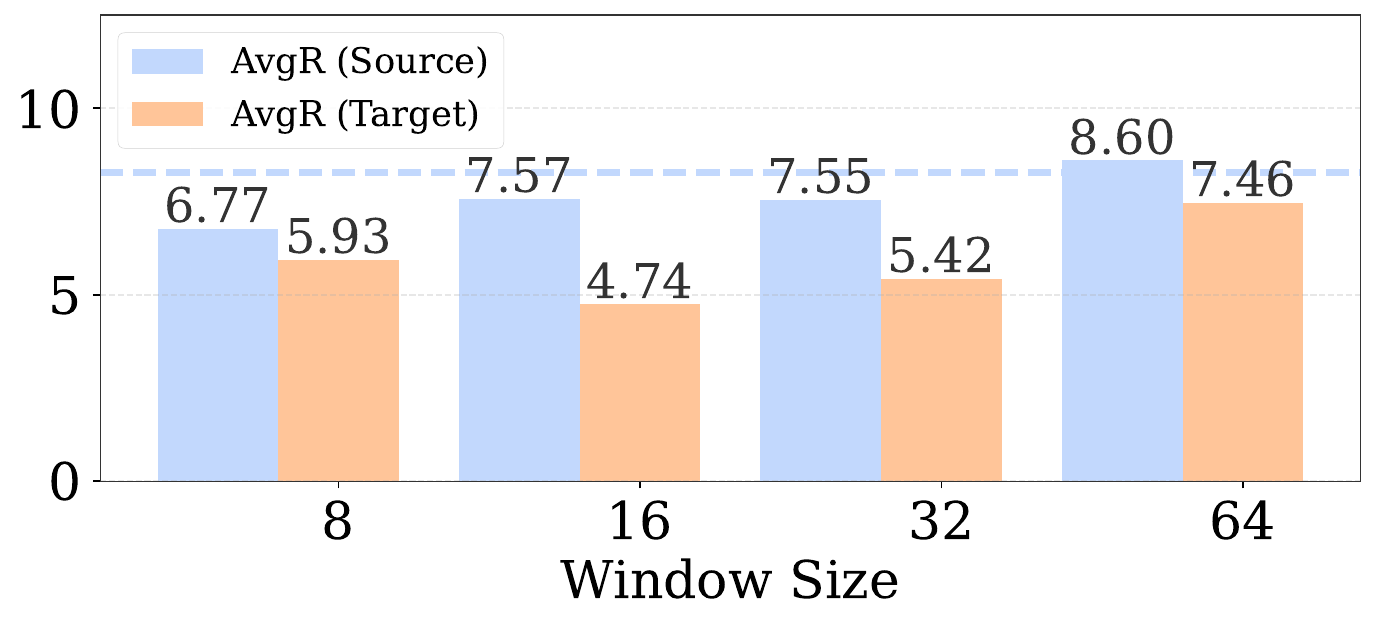}
        \caption{}
        \label{fig:ablation_c}
    \end{subfigure}%
    \begin{subfigure}{0.25\linewidth}
        \centering
        \includegraphics[width=\linewidth]{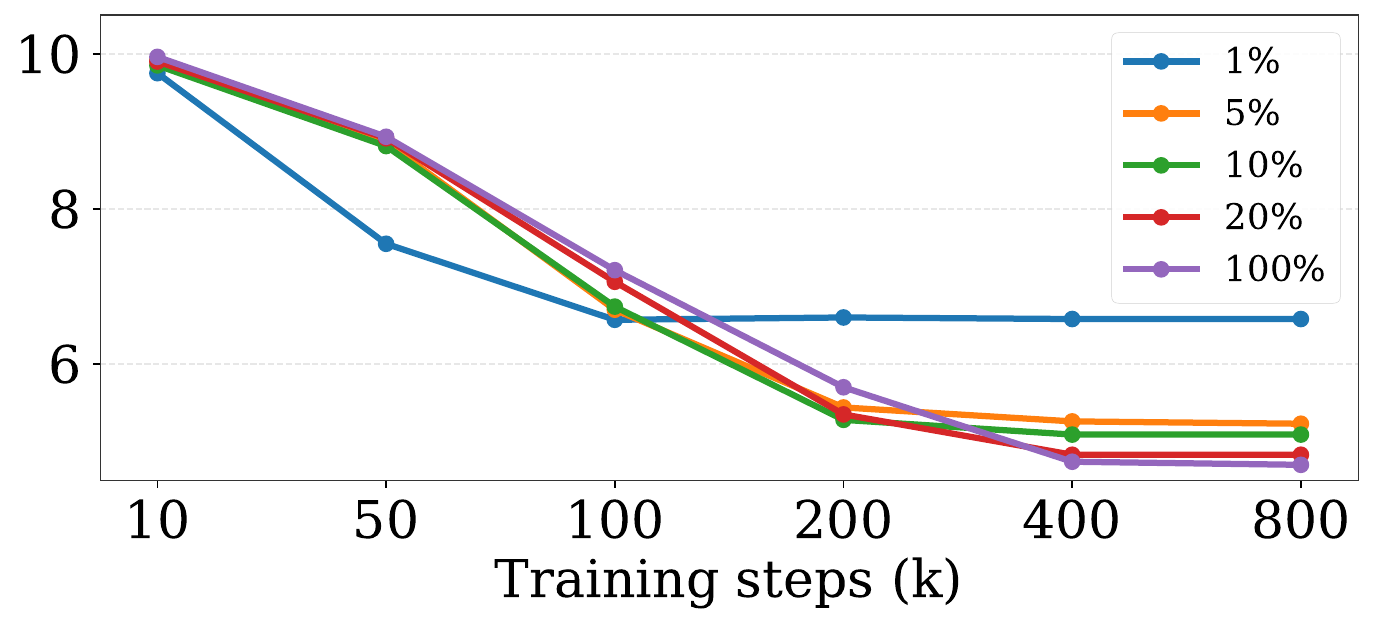}
        \caption{}
        \label{fig:ablation_d}
    \end{subfigure}%
    \\%
    \begin{subfigure}{0.25\linewidth}
        \centering
        \includegraphics[width=\linewidth]{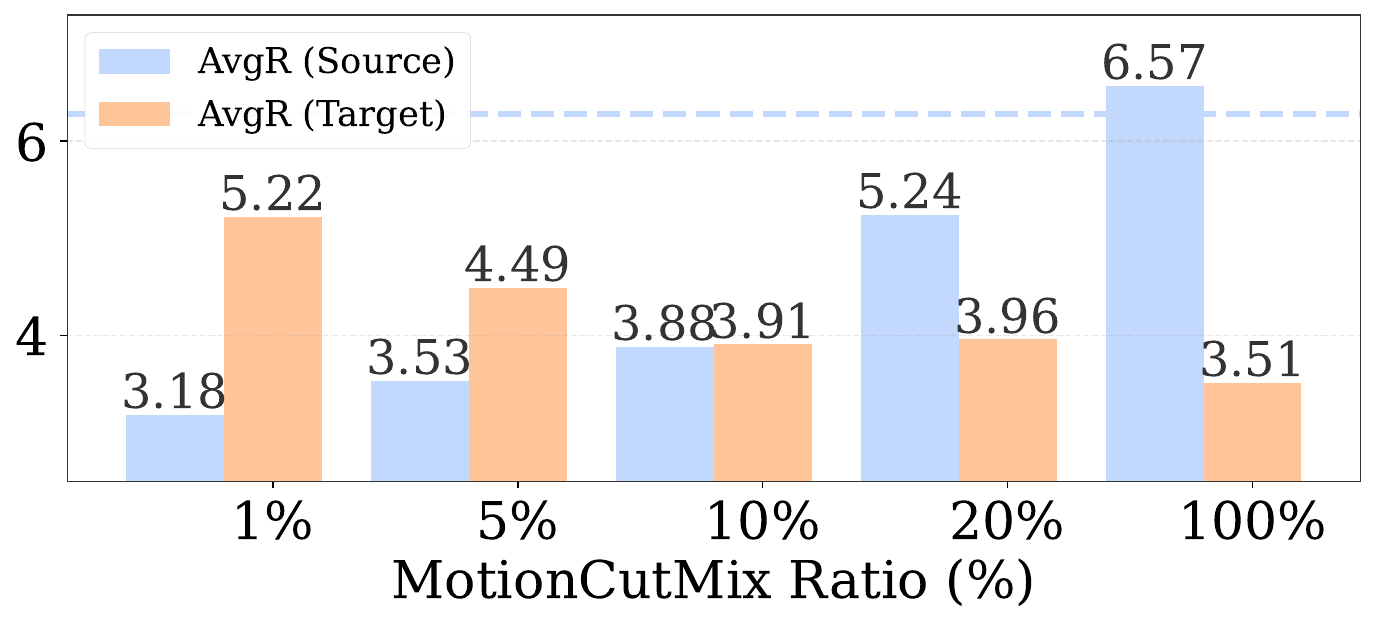}
        \caption{}
        \label{fig:ablation_e}
    \end{subfigure}%
    \begin{subfigure}{0.25\linewidth}
        \centering
        \includegraphics[width=\linewidth]{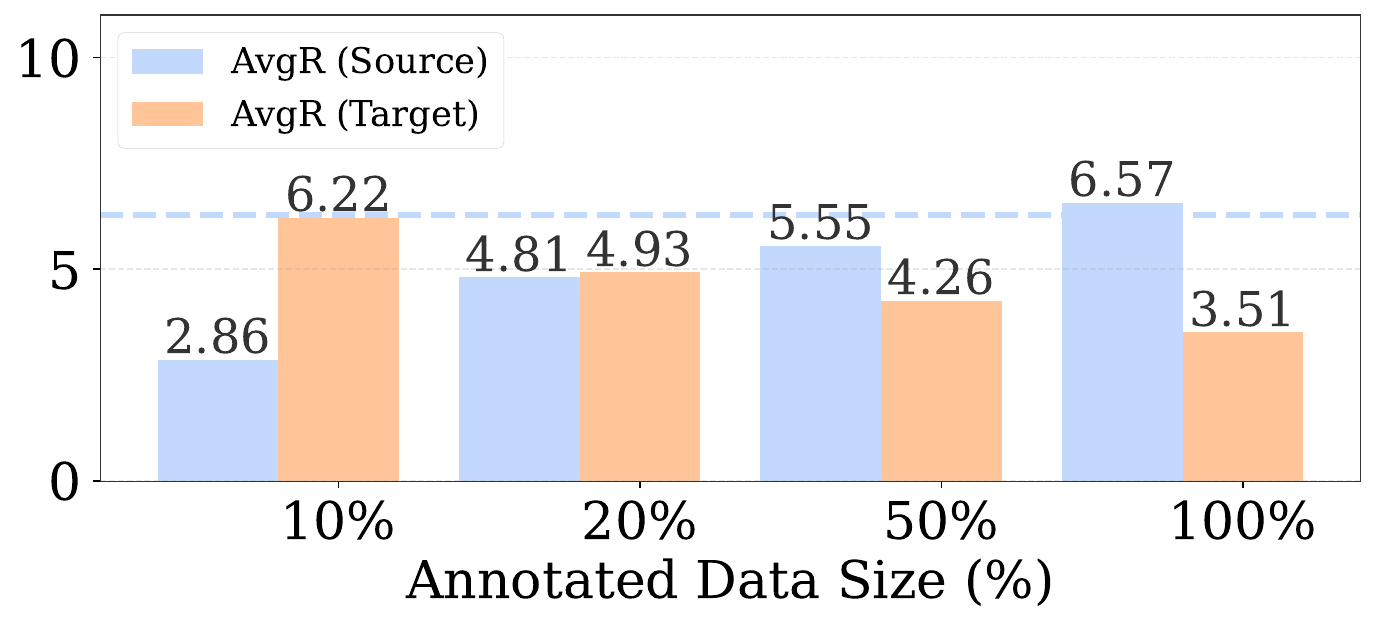}
        \caption{}
        \label{fig:ablation_f}
    \end{subfigure}%
    \begin{subfigure}{0.25\linewidth}
        \centering
        \includegraphics[width=\linewidth]{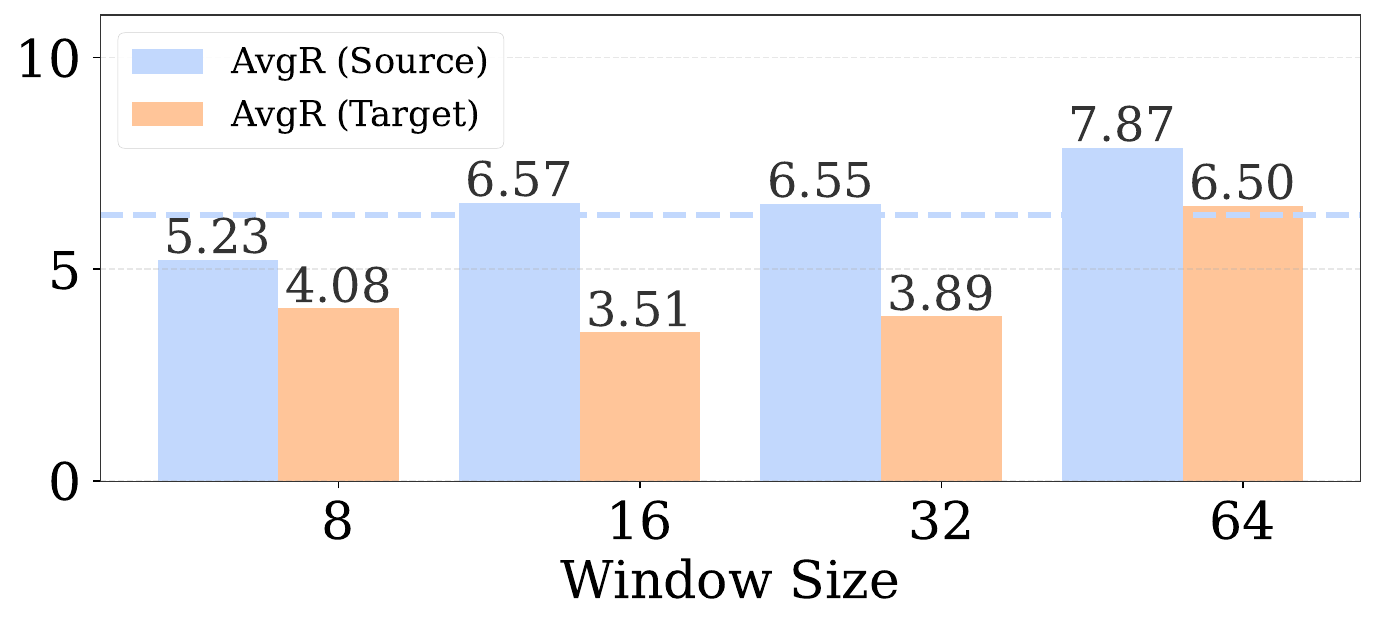}
        \caption{}
        \label{fig:ablation_g}
    \end{subfigure}%
    \begin{subfigure}{0.25\linewidth}
        \centering
        \includegraphics[width=\linewidth]{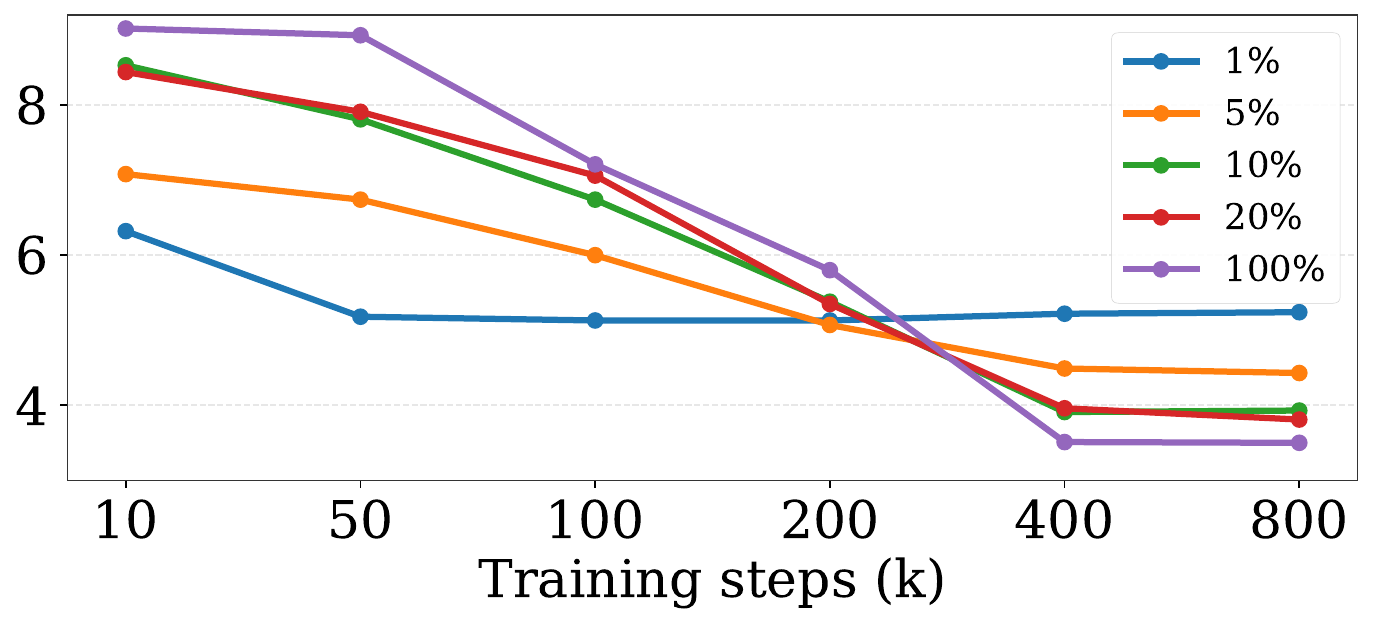}
        \caption{}%
        \label{fig:ablation_h}
    \end{subfigure}
    \caption{\textbf{Ablation analyses for body part replacement (a-d) and style transfer (e-h), reporting AvgR metrics.} Edited-to-Target AvgR shown only for (d) and (h), with blue dotted lines indicating real data Edited-to-Source AvgR. Parameters studied: (a,e) MotionCutMix ratio, (b,f) annotated data volume, (c,g) temporal window size, and (d,h) convergence patterns at varying MotionCutMix ratios. All training converges within 800k steps.}
    \label{fig:ablation}
    \vspace{-4pt}
\end{figure*}

\begin{figure}[ht!]
    \centering
    \includegraphics[width=\linewidth]{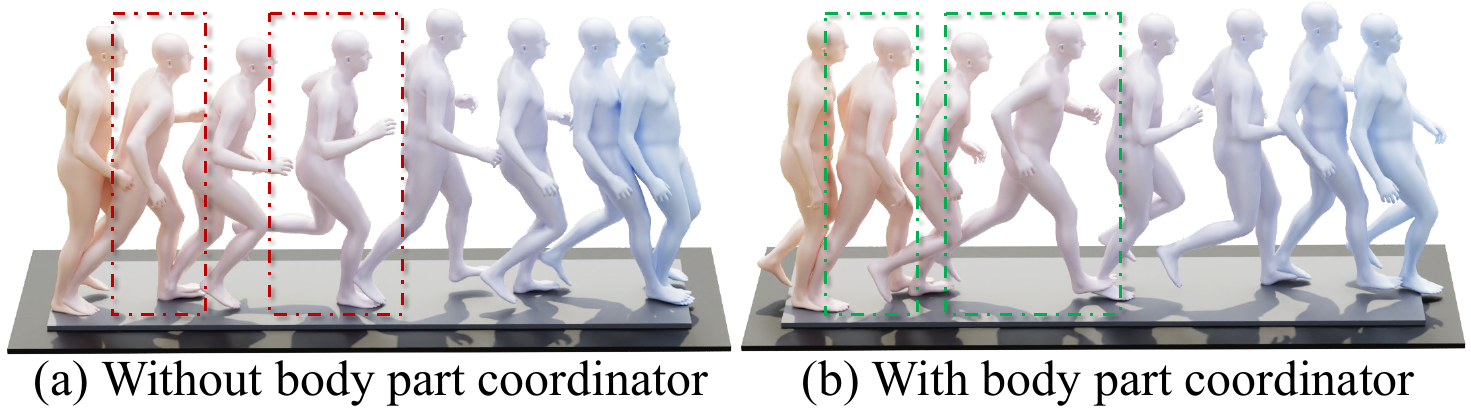}
    \caption{\textbf{Impact of body part coordinator on motion quality.} Examples show paired results using identical random seeds, highlighting how coordinator prevents unnatural synchronous movements of same-side limbs (arm and leg moving forward together).}
    \label{fig:disc}
\end{figure}

\subsection{Comparison Results}

Quantitative results in \cref{tab:results-main} demonstrate that our full method achieves superior performance across most metrics for both style and semantic edits. The retrieval scores indicate precise editing while preserving the original context. Qualitative results in \cref{fig:compare} showcase our approach's versatile editing capabilities. In semantic editing, our method successfully executes backward walking and crouching while maintaining upper body movements, whereas baseline methods fail to produce coherent motions. The style transfer examples highlight our method's sophisticated control, achieving pronounced style modifications while preserving the original motion's semantic content.

\begin{table}[b!]
    \centering
    \setlength{\tabcolsep}{1.5pt}
    \caption{\textbf{Quantitative comparison with TMED~\cite{athanasiou2024motionfix} evaluated on MotionFix dataset~\cite{athanasiou2024motionfix} using a gallery size of 32.} Results show means across 10 evaluation runs, with \textbf{bold} indicating best result.}
    \label{tab:tmed-mini-batch}
    \resizebox{\linewidth}{!}{%
        \begin{tabular}{lcccccccc}
            \toprule
            \multirow{2}{*}{Method} & \multicolumn{4}{c}{Edited-to-Source Retrieval} & \multicolumn{4}{c}{Edited-to-Target Retrieval} \\
            \cmidrule(r){2-5} \cmidrule(l){6-9} & R@1$\uparrow$ & R@2$\uparrow$ & R@3$\uparrow$ & AvgR$\downarrow$ & R@1$\uparrow$ & R@2$\uparrow$ & R@3$\uparrow$ & AvgR$\downarrow$ \\
            \midrule
            Real Data & 74.01 & 84.52 & 89.91 &  2.03 & 100.0 & 100.0 & 100.0 & 1.00 \\
            MDM-BP~\cite{tevet2022mdm} & 61.28 & 69.55 & 73.99 &  4.21 & 39.10 & 50.09 & 54.84 & 6.46 \\
            TMED~\cite{athanasiou2024motionfix} & 71.77 & 84.07 & 89.52 &  1.96 & 62.90 & 76.51 & 83.06 & 2.71 \\
            Ours w/o MCM & \textbf{83.47} & \textbf{90.42} & \textbf{92.84} &  \textbf{1.73} & \textbf{66.33} & \textbf{80.05} & \textbf{84.98} & \textbf{2.64} \\
            \bottomrule
        \end{tabular}
    }%
\end{table}

\cref{tab:tmed-mini-batch} presents batch-wise evaluation results on the MotionFix benchmark. Even without \strategy augmentation, our auto-regressive approach outperforms TMED and MDM-BP across all metrics. By processing long sequences through fixed-length windows, our method achieves both higher E2T scores for accurate editing and better E2R scores for context preservation. This demonstrates the effectiveness of our auto-regressive architecture over single-step approaches. Complete evaluation results at full-test set scale are provided in \cref{sec:detail-and-results-on-motionfix}.

\subsection{Ablation Results}

Our experiments show that \strategy significantly improves performance for both our method and TMED (\cref{tab:results-main}), demonstrating its broad applicability to motion editing tasks. The quality improvements from our body part coordinator are visible in \cref{fig:disc} and quantitatively supported by improved FID scores. Importantly, we find that merely learning from composed data is insufficient for proper body part coordination. Quantitative evaluation of guidance strength $\lambda$ and guidance steps count is presented in~\cref{sec:guidance}.

Our analysis through \cref{fig:ablation_a,fig:ablation_e} reveals that performance directly scales with the amount of augmented data---increasing the MotionCutMix Ratio leads to substantial gains in motion editing capabilities. When examining data efficiency in \cref{fig:ablation_b,fig:ablation_f}, we find that models with \strategy maintain strong performance even with reduced data scales compared to baseline models, indicating reduced dependence on annotated data volume. For temporal processing, our experiments in \cref{fig:ablation_c,fig:ablation_g} identify 16 frames as the optimal window size, effectively balancing data randomness with motion coherence. Training dynamics shown in \cref{fig:ablation_d,fig:ablation_h} demonstrate that despite introducing random variations, higher MotionCutMix ratios consistently improve performance without compromising training convergence.

\section{Conclusion}

This work introduces \model, a text-guided motion editing framework that enables precise modification of body parts and temporal segments while maintaining motion authenticity. We enhance the framework with \strategy for dynamic training augmentation and incorporate a body part coordinator for movement synchronization. Additionally, we contribute \dataset, a new MoCap and re-annotated dataset targeting three fundamental editing tasks: body part replacement, fine-grained adjustment, and style transfer.

Our work shows that for a specific motion editing task, minimal annotated data is sufficient. Moreover, by reducing the need for high-quality data (\eg MoCap data), our approach opens up broader applications. Specifically, we demonstrate that \model extends beyond motion editing to interactive modifications and complex compositional motion generation in \cref{sec:applications}.

\paragraph{Limitations}

Our approach exhibits limitations in processing long-term temporal dependencies and lacks spatial awareness for position-dependent instructions. A comprehensive discussion on limitations and future directions is provided in \cref{sec:limitations}.

\paragraph{Acknowledgment}

This work is supported in part by the National Natural Science Foundation of China (62376009), the Beijing Nova Program, the State Key Lab of General AI at Peking University, the PKU-BingJi Joint Laboratory for Artificial Intelligence, and the National Comprehensive Experimental Base for Governance of Intelligent Society, Wuhan East Lake High-Tech Development Zone.

\clearpage

{
    \small
    \bibliographystyle{ieeenat_fullname}
    \bibliography{reference_header,reference}

\begin{thebibliography}{72}
\providecommand{\natexlab}[1]{#1}
\providecommand{\url}[1]{\texttt{#1}}
\expandafter\ifx\csname urlstyle\endcsname\relax
  \providecommand{\doi}[1]{doi: #1}\else
  \providecommand{\doi}{doi: \begingroup \urlstyle{rm}\Url}\fi

\bibitem[Aberman et~al.(2020{\natexlab{a}})Aberman, Li, Lischinski, Sorkine-Hornung, Cohen-Or, and Chen]{aberman2020skeleton}
Kfir Aberman, Peizhuo Li, Dani Lischinski, Olga Sorkine-Hornung, Daniel Cohen-Or, and Baoquan Chen.
\newblock Skeleton-aware networks for deep motion retargeting.
\newblock \emph{ACM Transactions on Graphics (TOG)}, 39\penalty0 (4):\penalty0 62, 2020{\natexlab{a}}.

\bibitem[Aberman et~al.(2020{\natexlab{b}})Aberman, Weng, Lischinski, Cohen-Or, and Chen]{aberman2020unpaired}
Kfir Aberman, Yijia Weng, Dani Lischinski, Daniel Cohen-Or, and Baoquan Chen.
\newblock Unpaired motion style transfer from video to animation.
\newblock \emph{ACM Transactions on Graphics (TOG)}, 39\penalty0 (4):\penalty0 64, 2020{\natexlab{b}}.

\bibitem[Aliakbarian et~al.(2020)Aliakbarian, Saleh, Salzmann, Petersson, and Gould]{aliakbarian2020stochastic}
Sadegh Aliakbarian, Fatemeh~Sadat Saleh, Mathieu Salzmann, Lars Petersson, and Stephen Gould.
\newblock A stochastic conditioning scheme for diverse human motion prediction.
\newblock In \emph{Conference on Computer Vision and Pattern Recognition (CVPR)}, 2020.

\bibitem[Amaya et~al.(1996)Amaya, Bruderlin, and Calvert]{amaya1996emotion}
Kenji Amaya, Armin Bruderlin, and Tom Calvert.
\newblock Emotion from motion.
\newblock In \emph{Graphics interface}, 1996.

\bibitem[Athanasiou et~al.(2022)Athanasiou, Petrovich, Black, and Varol]{athanasiou2022teach}
Nikos Athanasiou, Mathis Petrovich, Michael~J Black, and G{\"u}l Varol.
\newblock Teach: Temporal action composition for 3d humans.
\newblock In \emph{International Conference on 3D Vision (3DV)}, 2022.

\bibitem[Athanasiou et~al.(2023)Athanasiou, Petrovich, Black, and Varol]{athanasiou2023sinc}
Nikos Athanasiou, Mathis Petrovich, Michael~J Black, and G{\"u}l Varol.
\newblock Sinc: Spatial composition of 3d human motions for simultaneous action generation.
\newblock In \emph{International Conference on Computer Vision (ICCV)}, 2023.

\bibitem[Athanasiou et~al.(2024)Athanasiou, Ceske, Diomataris, Black, and Varol]{athanasiou2024motionfix}
Nikos Athanasiou, Alp{\'a}r Ceske, Markos Diomataris, Michael~J Black, and G{\"u}l Varol.
\newblock Motionfix: Text-driven 3d human motion editing.
\newblock In \emph{ACM SIGGRAPH Conference Proceedings}, 2024.

\bibitem[Bie et~al.(2022)Bie, Guo, Leglaive, Girin, Moreno-Noguer, and Alameda-Pineda]{bie2022hit}
Xiaoyu Bie, Wen Guo, Simon Leglaive, Lauren Girin, Francesc Moreno-Noguer, and Xavier Alameda-Pineda.
\newblock Hit-dvae: Human motion generation via hierarchical transformer dynamical vae.
\newblock \emph{arXiv preprint arXiv:2204.01565}, 2022.

\bibitem[Brand and Hertzmann(2000)]{brand2000style}
Matthew Brand and Aaron Hertzmann.
\newblock Style machines.
\newblock In \emph{ACM SIGGRAPH Conference Proceedings}, 2000.

\bibitem[Brooks et~al.(2023)Brooks, Holynski, and Efros]{brooks2023instructpix2pix}
Tim Brooks, Aleksander Holynski, and Alexei~A Efros.
\newblock Instructpix2pix: Learning to follow image editing instructions.
\newblock In \emph{Conference on Computer Vision and Pattern Recognition (CVPR)}, 2023.

\bibitem[Chang et~al.(2022)Chang, Findlay, Zhang, and Shum]{chang2022unifying}
Ziyi Chang, Edmund J.~C. Findlay, Haozheng Zhang, and Hubert P.~H. Shum.
\newblock Unifying human motion synthesis and style transfer with denoising diffusion probabilistic models.
\newblock In \emph{Proceedings of International Conference on Computer Graphics Theory and Applications}, 2022.

\bibitem[Chen et~al.(2023)Chen, Jiang, Liu, Huang, Fu, Chen, and Yu]{chen2023executing}
Xin Chen, Biao Jiang, Wen Liu, Zilong Huang, Bin Fu, Tao Chen, and Gang Yu.
\newblock Executing your commands via motion diffusion in latent space.
\newblock In \emph{Conference on Computer Vision and Pattern Recognition (CVPR)}, 2023.

\bibitem[Dabral et~al.(2023)Dabral, Mughal, Golyanik, and Theobalt]{dabral2023mofusion}
Rishabh Dabral, Muhammad~Hamza Mughal, Vladislav Golyanik, and Christian Theobalt.
\newblock Mofusion: A framework for denoising-diffusion-based motion synthesis.
\newblock In \emph{Conference on Computer Vision and Pattern Recognition (CVPR)}, 2023.

\bibitem[Dai et~al.(2024)Dai, Chen, Wang, Liu, Dai, and Tang]{dai2024motionlcm}
Wenxun Dai, Ling-Hao Chen, Jingbo Wang, Jinpeng Liu, Bo Dai, and Yansong Tang.
\newblock Motionlcm: Real-time controllable motion generation via latent consistency model.
\newblock In \emph{European Conference on Computer Vision (ECCV)}, 2024.

\bibitem[Dong et~al.(2020)Dong, Aristidou, Shamir, Mahler, and Jain]{dong2020adult2child}
Yuzhu Dong, Andreas Aristidou, Ariel Shamir, Moshe Mahler, and Eakta Jain.
\newblock Adult2child: Motion style transfer using cyclegans.
\newblock In \emph{ACM SIGGRAPH Conference Proceedings}, 2020.

\bibitem[Gleicher(2001)]{gleicher2001motion}
Michael Gleicher.
\newblock Motion path editing.
\newblock In \emph{Proceedings of Symposium on Interactive 3D Graphics}, 2001.

\bibitem[Goel et~al.(2024)Goel, Wang, Liu, and Fatahalian]{goel2024iterative}
Purvi Goel, Kuan-Chieh Wang, C~Karen Liu, and Kayvon Fatahalian.
\newblock Iterative motion editing with natural language.
\newblock In \emph{ACM SIGGRAPH Conference Proceedings}, 2024.

\bibitem[Guo et~al.(2022)Guo, Zou, Zuo, Wang, Ji, Li, and Cheng]{guo2022humanml3d}
Chuan Guo, Shihao Zou, Xinxin Zuo, Sen Wang, Wei Ji, Xingyu Li, and Li Cheng.
\newblock Generating diverse and natural 3d human motions from text.
\newblock In \emph{Conference on Computer Vision and Pattern Recognition (CVPR)}, 2022.

\bibitem[Harvey et~al.(2020)Harvey, Yurick, Nowrouzezahrai, and Pal]{harvey2020robust}
F{\'e}lix~G Harvey, Mike Yurick, Derek Nowrouzezahrai, and Christopher Pal.
\newblock Robust motion in-betweening.
\newblock \emph{ACM Transactions on Graphics (TOG)}, 39\penalty0 (4):\penalty0 60, 2020.

\bibitem[Ho and Salimans(2022)]{ho2022classifier}
Jonathan Ho and Tim Salimans.
\newblock Classifier-free diffusion guidance.
\newblock \emph{arXiv preprint arXiv:2207.12598}, 2022.

\bibitem[Ho et~al.(2020)Ho, Jain, and Abbeel]{ho2020denoising}
Jonathan Ho, Ajay Jain, and Pieter Abbeel.
\newblock Denoising diffusion probabilistic models.
\newblock In \emph{Advances in Neural Information Processing Systems (NeurIPS)}, 2020.

\bibitem[Holden et~al.(2016)Holden, Saito, and Komura]{holden2016deep}
Daniel Holden, Jun Saito, and Taku Komura.
\newblock A deep learning framework for character motion synthesis and editing.
\newblock \emph{ACM Transactions on Graphics (TOG)}, 35\penalty0 (4):\penalty0 138, 2016.

\bibitem[Hong et~al.(2022)Hong, Zhang, Pan, Cai, Yang, and Liu]{hong2022avatarclip}
Fangzhou Hong, Mingyuan Zhang, Liang Pan, Zhongang Cai, Lei Yang, and Ziwei Liu.
\newblock Avatarclip: zero-shot text-driven generation and animation of 3d avatars.
\newblock \emph{ACM Transactions on Graphics (TOG)}, 41\penalty0 (4):\penalty0 161, 2022.

\bibitem[Hsu et~al.(2005)Hsu, Pulli, and Popovi\'{c}]{hsu2005style}
Eugene Hsu, Kari Pulli, and Jovan Popovi\'{c}.
\newblock Style translation for human motion.
\newblock \emph{ACM Transactions on Graphics (TOG)}, 24\penalty0 (3):\penalty0 1082, 2005.

\bibitem[Huang et~al.(2022)Huang, Mo, Liang, and Gao]{huang2022unpaired}
Yue Huang, Haoran Mo, Xiao Liang, and Chengying Gao.
\newblock Unpaired motion style transfer with motion-oriented projection flow network.
\newblock In \emph{International Conference on Multimedia and Expo (ICME)}, 2022.

\bibitem[Huang et~al.(2025)Huang, Wan, Yang, Callison-Burch, Yatskar, and Liu]{huang2024como}
Yiming Huang, Weilin Wan, Yue Yang, Chris Callison-Burch, Mark Yatskar, and Lingjie Liu.
\newblock Como: Controllable motion generation through language guided pose code editing.
\newblock In \emph{European Conference on Computer Vision (ECCV)}, 2025.

\bibitem[Ikemoto et~al.(2009)Ikemoto, Arikan, and Forsyth]{ikemoto2009generalizing}
Leslie Ikemoto, Okan Arikan, and David Forsyth.
\newblock Generalizing motion edits with gaussian processes.
\newblock \emph{ACM Transactions on Graphics (TOG)}, 28\penalty0 (1):\penalty0 1, 2009.

\bibitem[Jang et~al.(2022)Jang, Park, and Lee]{jang2022motion}
Deok-Kyeong Jang, Soomin Park, and Sung-Hee Lee.
\newblock Motion puzzle: Arbitrary motion style transfer by body part.
\newblock \emph{ACM Transactions on Graphics (TOG)}, 41\penalty0 (3):\penalty0 33, 2022.

\bibitem[Jiang et~al.(2023)Jiang, Chen, Liu, Yu, Yu, and Chen]{jiang2023motiongpt}
Biao Jiang, Xin Chen, Wen Liu, Jingyi Yu, Gang Yu, and Tao Chen.
\newblock Motiongpt: Human motion as a foreign language.
\newblock In \emph{Advances in Neural Information Processing Systems (NeurIPS)}, 2023.

\bibitem[Jiang et~al.(2024)Jiang, Zhang, Li, Ma, Wang, Chen, Liu, Zhu, and Huang]{jiang2024scaling}
Nan Jiang, Zhiyuan Zhang, Hongjie Li, Xiaoxuan Ma, Zan Wang, Yixin Chen, Tengyu Liu, Yixin Zhu, and Siyuan Huang.
\newblock Scaling up dynamic human-scene interaction modeling.
\newblock In \emph{Conference on Computer Vision and Pattern Recognition (CVPR)}, 2024.

\bibitem[Jin et~al.(2024)Jin, Wu, Fan, Sun, Yang, and Yuan]{jin2024act}
Peng Jin, Yang Wu, Yanbo Fan, Zhongqian Sun, Wei Yang, and Li Yuan.
\newblock Act as you wish: Fine-grained control of motion diffusion model with hierarchical semantic graphs.
\newblock In \emph{Advances in Neural Information Processing Systems (NeurIPS)}, 2024.

\bibitem[Kim et~al.(2023)Kim, Kim, and Choi]{kim2023flame}
Jihoon Kim, Jiseob Kim, and Sungjoon Choi.
\newblock Flame: Free-form language-based motion synthesis \& editing.
\newblock In \emph{AAAI Conference on Artificial Intelligence (AAAI)}, 2023.

\bibitem[Kim et~al.(2009)Kim, Hyun, Kim, and Lee]{kim2009synchronized}
Manmyung Kim, Kyunglyul Hyun, Jongmin Kim, and Jehee Lee.
\newblock Synchronized multi-character motion editing.
\newblock \emph{ACM Transactions on Graphics (TOG)}, 28\penalty0 (3):\penalty0 79, 2009.

\bibitem[Kingma and Ba(2014)]{kingma2014adam}
Diederik~P Kingma and Jimmy Ba.
\newblock Adam: A method for stochastic optimization.
\newblock \emph{arXiv preprint arXiv:1412.6980}, 2014.

\bibitem[Lin and Amer(2018)]{lin2018human}
Xiao Lin and Mohamed~R Amer.
\newblock Human motion modeling using dvgans.
\newblock \emph{arXiv preprint arXiv:1804.10652}, 2018.

\bibitem[Lockwood and Singh(2011)]{lockwood2011biomechanically}
Noah Lockwood and Karan Singh.
\newblock Biomechanically-inspired motion path editing.
\newblock In \emph{ACM SIGGRAPH / Eurographics Symposium on Computer Animation (SCA)}, 2011.

\bibitem[Loper et~al.(2015)Loper, Mahmood, Romero, Pons-Moll, and Black]{loper2015smpl}
Matthew Loper, Naureen Mahmood, Javier Romero, Gerard Pons-Moll, and Michael~J Black.
\newblock Smpl: A skinned multi-person linear model.
\newblock \emph{ACM Transactions on Graphics (TOG)}, 34\penalty0 (6):\penalty0 248, 2015.

\bibitem[Loshchilov(2017)]{loshchilov2017decoupled}
I Loshchilov.
\newblock Decoupled weight decay regularization.
\newblock \emph{arXiv preprint arXiv:1711.05101}, 2017.

\bibitem[Ma et~al.(2010)Ma, Xia, Hodgins, Yang, Li, and Wang]{ma2010modeling}
Wanli Ma, Shihong Xia, Jessica~K Hodgins, Xiao Yang, Chunpeng Li, and Zhaoqi Wang.
\newblock Modeling style and variation in human motion.
\newblock In \emph{ACM SIGGRAPH / Eurographics Symposium on Computer Animation (SCA)}, 2010.

\bibitem[Mahmood et~al.(2019)Mahmood, Ghorbani, Troje, Pons-Moll, and Black]{mahmood2019amass}
Naureen Mahmood, Nima Ghorbani, Nikolaus~F Troje, Gerard Pons-Moll, and Michael~J Black.
\newblock Amass: Archive of motion capture as surface shapes.
\newblock In \emph{International Conference on Computer Vision (ICCV)}, 2019.

\bibitem[Pavlakos et~al.(2019)Pavlakos, Choutas, Ghorbani, Bolkart, Osman, Tzionas, and Black]{pavlakos2019expressive}
Georgios Pavlakos, Vasileios Choutas, Nima Ghorbani, Timo Bolkart, Ahmed~AA Osman, Dimitrios Tzionas, and Michael~J Black.
\newblock Expressive body capture: 3d hands, face, and body from a single image.
\newblock In \emph{Conference on Computer Vision and Pattern Recognition (CVPR)}, 2019.

\bibitem[Petrovich et~al.(2022)Petrovich, Black, and Varol]{petrovich2022temos}
Mathis Petrovich, Michael~J Black, and G{\"u}l Varol.
\newblock Temos: Generating diverse human motions from textual descriptions.
\newblock In \emph{European Conference on Computer Vision (ECCV)}, 2022.

\bibitem[Petrovich et~al.(2023)Petrovich, Black, and Varol]{petrovich2023tmr}
Mathis Petrovich, Michael~J Black, and G{\"u}l Varol.
\newblock Tmr: Text-to-motion retrieval using contrastive 3d human motion synthesis.
\newblock In \emph{International Conference on Computer Vision (ICCV)}, 2023.

\bibitem[Petrovich et~al.(2024)Petrovich, Litany, Iqbal, Black, Varol, Bin~Peng, and Rempe]{petrovich2024multi}
Mathis Petrovich, Or Litany, Umar Iqbal, Michael~J Black, Gul Varol, Xue Bin~Peng, and Davis Rempe.
\newblock Multi-track timeline control for text-driven 3d human motion generation.
\newblock In \emph{Conference on Computer Vision and Pattern Recognition (CVPR)}, 2024.

\bibitem[Pinyoanuntapong et~al.(2024)Pinyoanuntapong, Wang, Lee, and Chen]{pinyoanuntapong2024mmm}
Ekkasit Pinyoanuntapong, Pu Wang, Minwoo Lee, and Chen Chen.
\newblock Mmm: Generative masked motion model.
\newblock In \emph{Conference on Computer Vision and Pattern Recognition (CVPR)}, 2024.

\bibitem[Punnakkal et~al.(2021)Punnakkal, Chandrasekaran, Athanasiou, Quiros-Ramirez, and Black]{punnakkal2021babel}
Abhinanda~R Punnakkal, Arjun Chandrasekaran, Nikos Athanasiou, Alejandra Quiros-Ramirez, and Michael~J Black.
\newblock Babel: Bodies, action and behavior with english labels.
\newblock In \emph{Conference on Computer Vision and Pattern Recognition (CVPR)}, 2021.

\bibitem[Qian et~al.(2024)Qian, Xiao, Wu, Yang, Li, Wang, Wang, Kou, and Zhang]{qian2024smcd}
Ziyun Qian, Zeyu Xiao, Zhenyi Wu, Dingkang Yang, Mingcheng Li, Shunli Wang, Shuaibing Wang, Dongliang Kou, and Lihua Zhang.
\newblock Smcd: High realism motion style transfer via mamba-based diffusion.
\newblock \emph{arXiv preprint arXiv:2405.02844}, 2024.

\bibitem[Radford et~al.(2021)Radford, Kim, Hallacy, Ramesh, Goh, Agarwal, Sastry, Askell, Mishkin, Clark, et~al.]{radford2021learning}
Alec Radford, Jong~Wook Kim, Chris Hallacy, Aditya Ramesh, Gabriel Goh, Sandhini Agarwal, Girish Sastry, Amanda Askell, Pamela Mishkin, Jack Clark, et~al.
\newblock Learning transferable visual models from natural language supervision.
\newblock In \emph{International Conference on Machine Learning (ICML)}, 2021.

\bibitem[Shafir et~al.(2024)Shafir, Tevet, Kapon, and Bermano]{shafir2024human}
Yoni Shafir, Guy Tevet, Roy Kapon, and Amit~Haim Bermano.
\newblock Human motion diffusion as a generative prior.
\newblock In \emph{International Conference on Learning Representations (ICLR)}, 2024.

\bibitem[Shahroudy et~al.(2016)Shahroudy, Liu, Ng, and Wang]{shahroudy2016ntu}
Amir Shahroudy, Jun Liu, Tian-Tsong Ng, and Gang Wang.
\newblock Ntu rgb+ d: A large scale dataset for 3d human activity analysis.
\newblock In \emph{Conference on Computer Vision and Pattern Recognition (CVPR)}, 2016.

\bibitem[Shi et~al.(2024)Shi, Wang, Jiang, Lin, Dai, and Peng]{shi2024interactive}
Yi Shi, Jingbo Wang, Xuekun Jiang, Bingkun Lin, Bo Dai, and Xue~Bin Peng.
\newblock Interactive character control with auto-regressive motion diffusion models.
\newblock \emph{ACM Transactions on Graphics (TOG)}, 43\penalty0 (4):\penalty0 143, 2024.

\bibitem[Sohl-Dickstein et~al.(2015)Sohl-Dickstein, Weiss, Maheswaranathan, and Ganguli]{sohl2015deep}
Jascha Sohl-Dickstein, Eric Weiss, Niru Maheswaranathan, and Surya Ganguli.
\newblock Deep unsupervised learning using nonequilibrium thermodynamics.
\newblock In \emph{International Conference on Machine Learning (ICML)}, 2015.

\bibitem[Sun et~al.(2024)Sun, Zheng, Huang, Ma, Huang, and Hu]{sun2024lgtm}
Haowen Sun, Ruikun Zheng, Haibin Huang, Chongyang Ma, Hui Huang, and Ruizhen Hu.
\newblock Lgtm: Local-to-global text-driven human motion diffusion model.
\newblock In \emph{ACM SIGGRAPH Conference Proceedings}, 2024.

\bibitem[Tao et~al.(2022)Tao, Zhan, Chen, and van~de Panne]{tao2022style}
Tianxin Tao, Xiaohang Zhan, Zhongquan Chen, and Michiel van~de Panne.
\newblock Style-erd: Responsive and coherent online motion style transfer.
\newblock In \emph{Conference on Computer Vision and Pattern Recognition (CVPR)}, 2022.

\bibitem[Tevet et~al.(2022{\natexlab{a}})Tevet, Gordon, Hertz, Bermano, and Cohen-Or]{tevet2022motionclip}
Guy Tevet, Brian Gordon, Amir Hertz, Amit~H Bermano, and Daniel Cohen-Or.
\newblock Motionclip: Exposing human motion generation to clip space.
\newblock In \emph{European Conference on Computer Vision (ECCV)}, 2022{\natexlab{a}}.

\bibitem[Tevet et~al.(2022{\natexlab{b}})Tevet, Raab, Gordon, Shafir, Cohen-or, and Bermano]{tevet2022mdm}
Guy Tevet, Sigal Raab, Brian Gordon, Yoni Shafir, Daniel Cohen-or, and Amit~Haim Bermano.
\newblock Human motion diffusion model.
\newblock In \emph{International Conference on Learning Representations (ICLR)}, 2022{\natexlab{b}}.

\bibitem[Tseng et~al.(2023)Tseng, Castellon, and Liu]{tseng2023edge}
Jonathan Tseng, Rodrigo Castellon, and Karen Liu.
\newblock Edge: Editable dance generation from music.
\newblock In \emph{Conference on Computer Vision and Pattern Recognition (CVPR)}, 2023.

\bibitem[Unuma et~al.(1995)Unuma, Anjyo, and Takeuchi]{unuma1995fourier}
Munetoshi Unuma, Ken Anjyo, and Ryozo Takeuchi.
\newblock Fourier principles for emotion-based human figure animation.
\newblock In \emph{ACM SIGGRAPH Conference Proceedings}, 1995.

\bibitem[Vaswani(2017)]{vaswani2017attention}
A Vaswani.
\newblock Attention is all you need.
\newblock In \emph{Advances in Neural Information Processing Systems (NeurIPS)}, 2017.

\bibitem[Wang et~al.(2007)Wang, Fleet, and Hertzmann]{wang2007multifactor}
Jack~M Wang, David~J Fleet, and Aaron Hertzmann.
\newblock Multifactor gaussian process models for style-content separation.
\newblock In \emph{International Conference on Machine Learning (ICML)}, 2007.

\bibitem[Wang et~al.(2020)Wang, Yu, Zhao, Zhang, Zhou, Yuan, and Chen]{wang2020learning}
Zhenyi Wang, Ping Yu, Yang Zhao, Ruiyi Zhang, Yufan Zhou, Junsong Yuan, and Changyou Chen.
\newblock Learning diverse stochastic human-action generators by learning smooth latent transitions.
\newblock In \emph{AAAI Conference on Artificial Intelligence (AAAI)}, 2020.

\bibitem[Witkin and Popovic(1995)]{witkin1995motion}
Andrew Witkin and Zoran Popovic.
\newblock Motion warping.
\newblock In \emph{ACM SIGGRAPH Conference Proceedings}, 1995.

\bibitem[Xia et~al.(2015)Xia, Wang, Chai, and Hodgins]{xia2015realtime}
Shihong Xia, Congyi Wang, Jinxiang Chai, and Jessica Hodgins.
\newblock Realtime style transfer for unlabeled heterogeneous human motion.
\newblock \emph{ACM Transactions on Graphics (TOG)}, 34\penalty0 (4):\penalty0 119, 2015.

\bibitem[Xie et~al.(2023)Xie, Jampani, Zhong, Sun, and Jiang]{xie2023omnicontrol}
Yiming Xie, Varun Jampani, Lei Zhong, Deqing Sun, and Huaizu Jiang.
\newblock Omnicontrol: Control any joint at any time for human motion generation.
\newblock In \emph{International Conference on Learning Representations (ICLR)}, 2023.

\bibitem[Xu et~al.(2023)Xu, Li, Wang, and Gui]{xu2023interdiff}
Sirui Xu, Zhengyuan Li, Yu-Xiong Wang, and Liang-Yan Gui.
\newblock Interdiff: Generating 3d human-object interactions with physics-informed diffusion.
\newblock In \emph{International Conference on Computer Vision (ICCV)}, 2023.

\bibitem[Yan et~al.(2018)Yan, Rastogi, Villegas, Sunkavalli, Shechtman, Hadap, Yumer, and Lee]{yan2018mt}
Xinchen Yan, Akash Rastogi, Ruben Villegas, Kalyan Sunkavalli, Eli Shechtman, Sunil Hadap, Ersin Yumer, and Honglak Lee.
\newblock Mt-vae: Learning motion transformations to generate multimodal human dynamics.
\newblock In \emph{European Conference on Computer Vision (ECCV)}, 2018.

\bibitem[Yin et~al.(2024)Yin, Yu, Yin, Kragic, and Bj{\"o}rkman]{yin2024scalable}
Wenjie Yin, Yi Yu, Hang Yin, Danica Kragic, and M{\aa}rten Bj{\"o}rkman.
\newblock Scalable motion style transfer with constrained diffusion generation.
\newblock In \emph{AAAI Conference on Artificial Intelligence (AAAI)}, 2024.

\bibitem[Yun et~al.(2019)Yun, Han, Oh, Chun, Choe, and Yoo]{yun2019cutmix}
Sangdoo Yun, Dongyoon Han, Seong~Joon Oh, Sanghyuk Chun, Junsuk Choe, and Youngjoon Yoo.
\newblock Cutmix: Regularization strategy to train strong classifiers with localizable features.
\newblock In \emph{International Conference on Computer Vision (ICCV)}, 2019.

\bibitem[Zhang et~al.(2024{\natexlab{a}})Zhang, Cai, Pan, Hong, Guo, Yang, and Liu]{zhang2024motiondiffuse}
Mingyuan Zhang, Zhongang Cai, Liang Pan, Fangzhou Hong, Xinying Guo, Lei Yang, and Ziwei Liu.
\newblock Motiondiffuse: Text-driven human motion generation with diffusion model.
\newblock \emph{Transactions on Pattern Analysis and Machine Intelligence (TPAMI)}, 46\penalty0 (6):\penalty0 4115--4128, 2024{\natexlab{a}}.

\bibitem[Zhang et~al.(2024{\natexlab{b}})Zhang, Li, Cai, Ren, Yang, and Liu]{zhang2024finemogen}
Mingyuan Zhang, Huirong Li, Zhongang Cai, Jiawei Ren, Lei Yang, and Ziwei Liu.
\newblock Finemogen: Fine-grained spatio-temporal motion generation and editing.
\newblock In \emph{Advances in Neural Information Processing Systems (NeurIPS)}, 2024{\natexlab{b}}.

\bibitem[Zhao et~al.(2023)Zhao, Zhang, Wang, Beeler, and Tang]{zhao2023synthesizing}
Kaifeng Zhao, Yan Zhang, Shaofei Wang, Thabo Beeler, and Siyu Tang.
\newblock Synthesizing diverse human motions in 3d indoor scenes.
\newblock In \emph{International Conference on Computer Vision (ICCV)}, 2023.

\bibitem[Zhong et~al.(2023)Zhong, Hu, Zhang, and Xia]{zhong2023attt2m}
Chongyang Zhong, Lei Hu, Zihao Zhang, and Shihong Xia.
\newblock Attt2m: Text-driven human motion generation with multi-perspective attention mechanism.
\newblock In \emph{International Conference on Computer Vision (ICCV)}, 2023.

\end{thebibliography}
}

\clearpage
\appendix
\renewcommand\thefigure{A\arabic{figure}}
\setcounter{figure}{0}
\renewcommand\thetable{A\arabic{table}}
\setcounter{table}{0}
\renewcommand\theequation{A\arabic{equation}}
\setcounter{equation}{0}
\pagenumbering{arabic}
\renewcommand*{\thepage}{A\arabic{page}}
\setcounter{footnote}{0}

\section{Additional Qualitative Results}\label{sec:additional-results}

\begin{figure*}[ht!]
    \centering
    \includegraphics[width=0.9\textwidth]{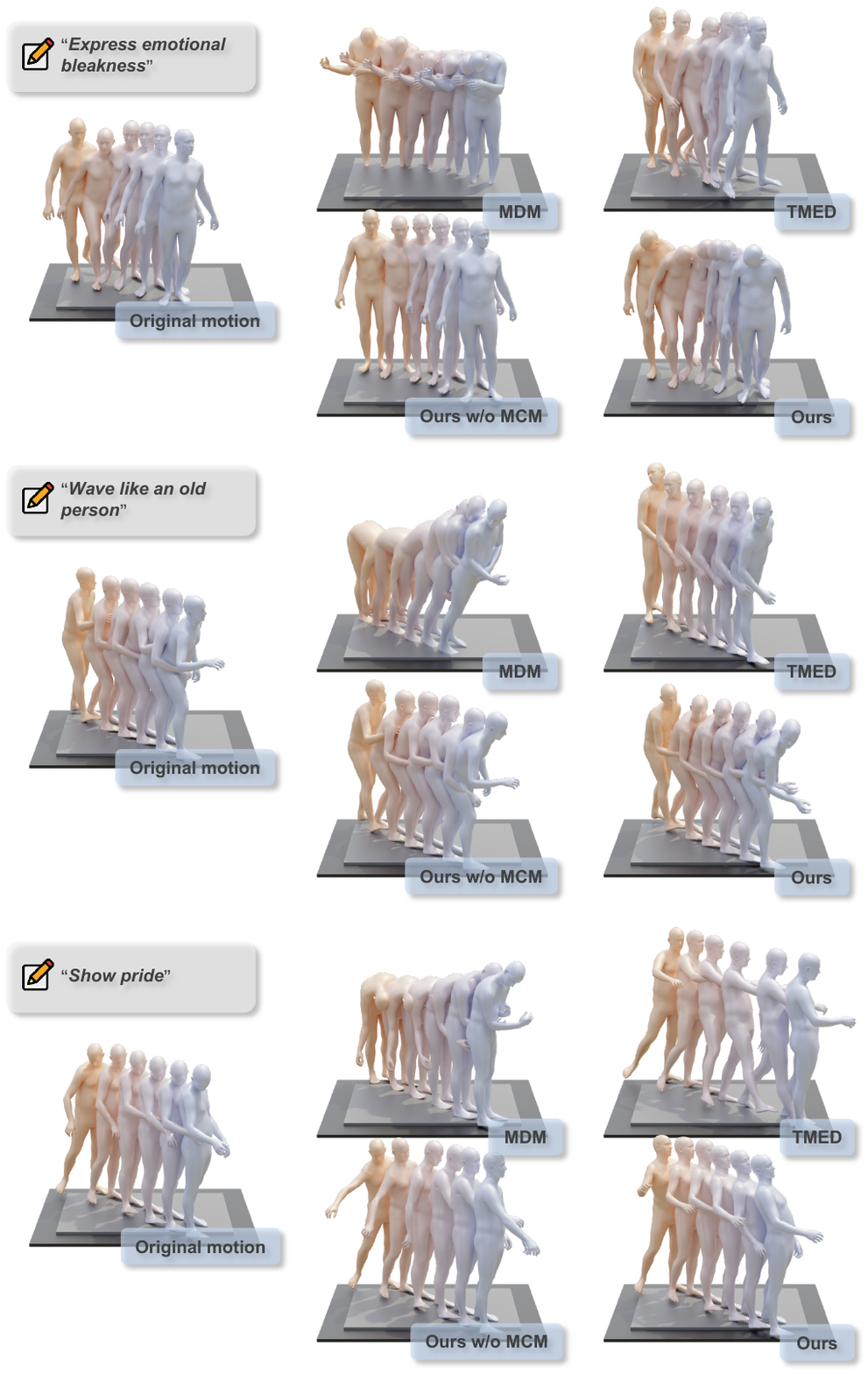}
    \caption{\textbf{Comparison with baselines and ablations on style transfer.}}
    \label{fig:st_1}
\end{figure*}

\begin{figure*}[ht!]
    \centering
    \includegraphics[width=0.9\textwidth]{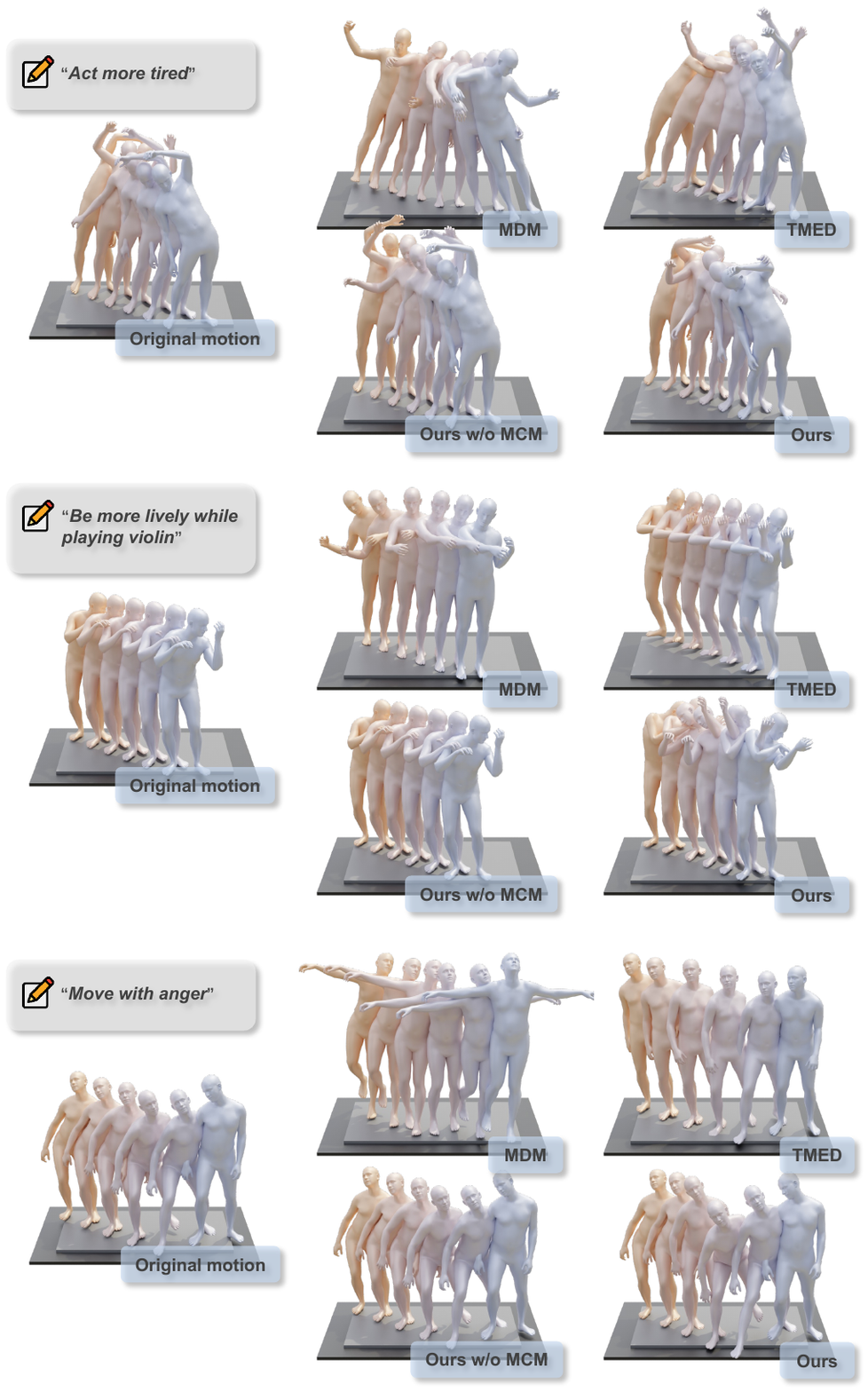}
    \caption{\textbf{Comparison with baselines and ablations on style transfer.}}
    \label{fig:st_2}
\end{figure*}

\begin{figure*}[ht!]
    \centering
    \includegraphics[width=0.9\textwidth]{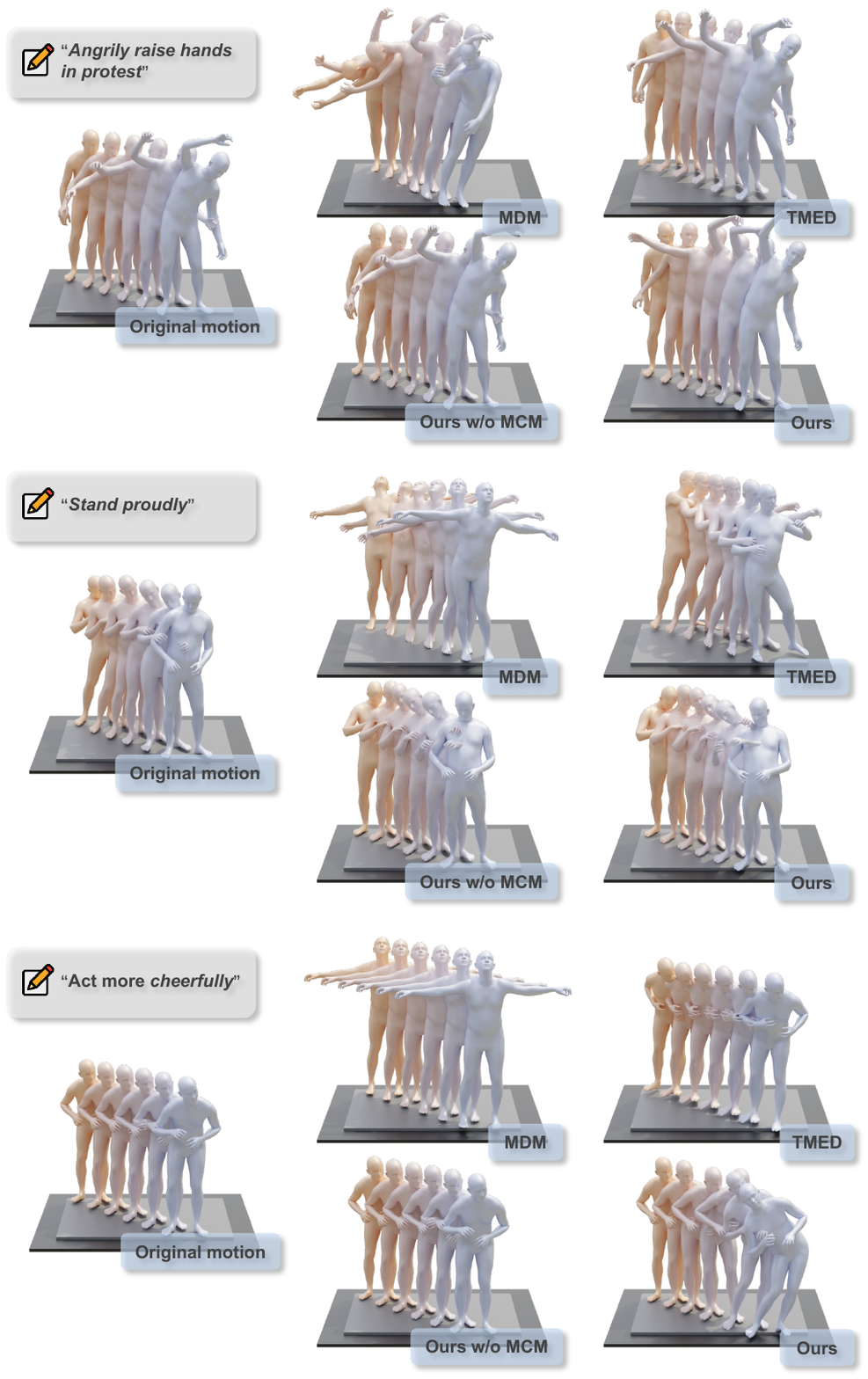}
    \caption{\textbf{Comparison with baselines and ablations on style transfer.}}
    \label{fig:st_3}
\end{figure*}

\begin{figure*}[ht!]
    \centering
    \includegraphics[width=0.9\textwidth]{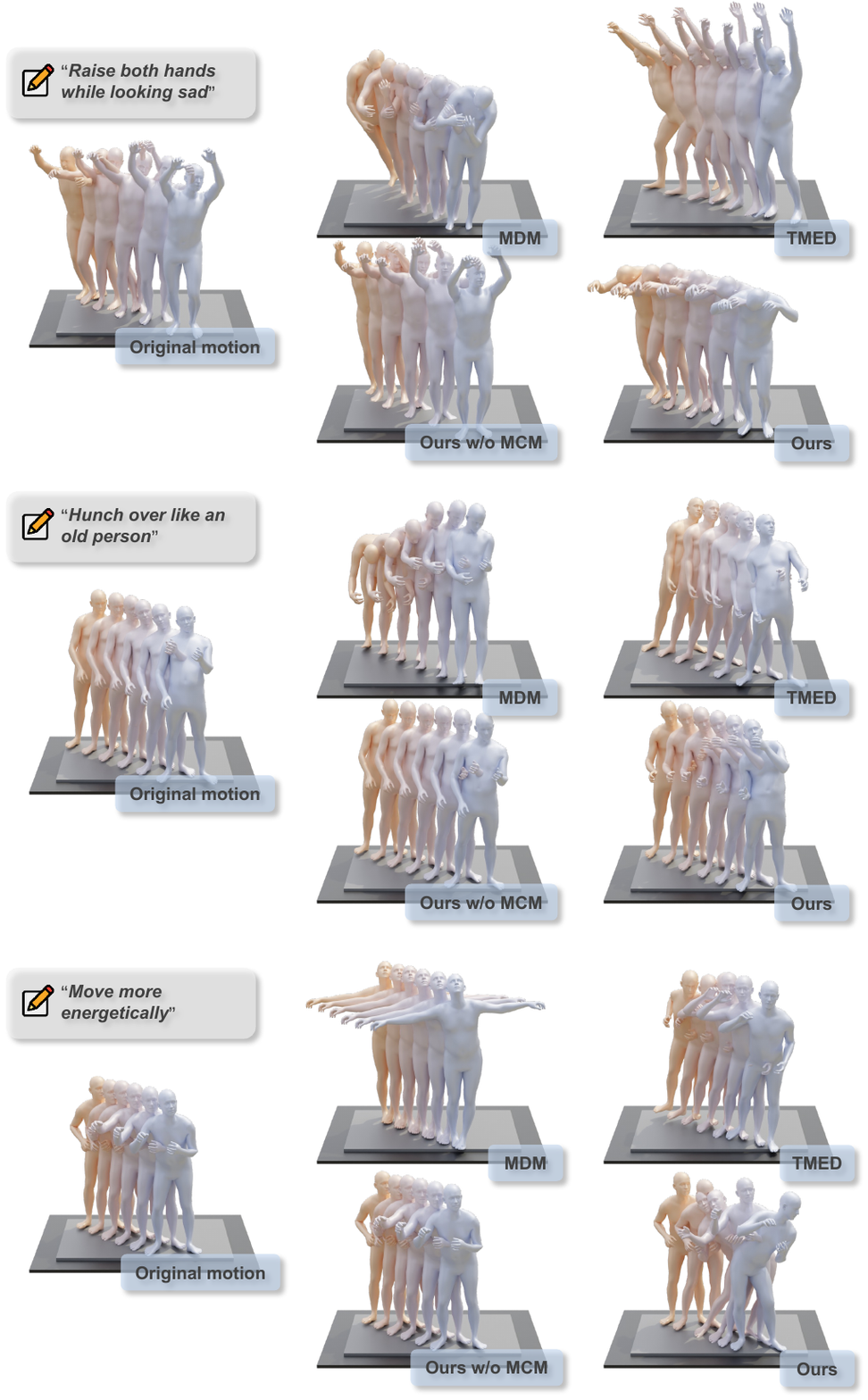}
    \caption{\textbf{Comparison with baselines and ablations on style transfer.}}
    \label{fig:st_4}
\end{figure*}

\begin{figure*}[ht!]
    \centering
    \includegraphics[width=0.9\textwidth]{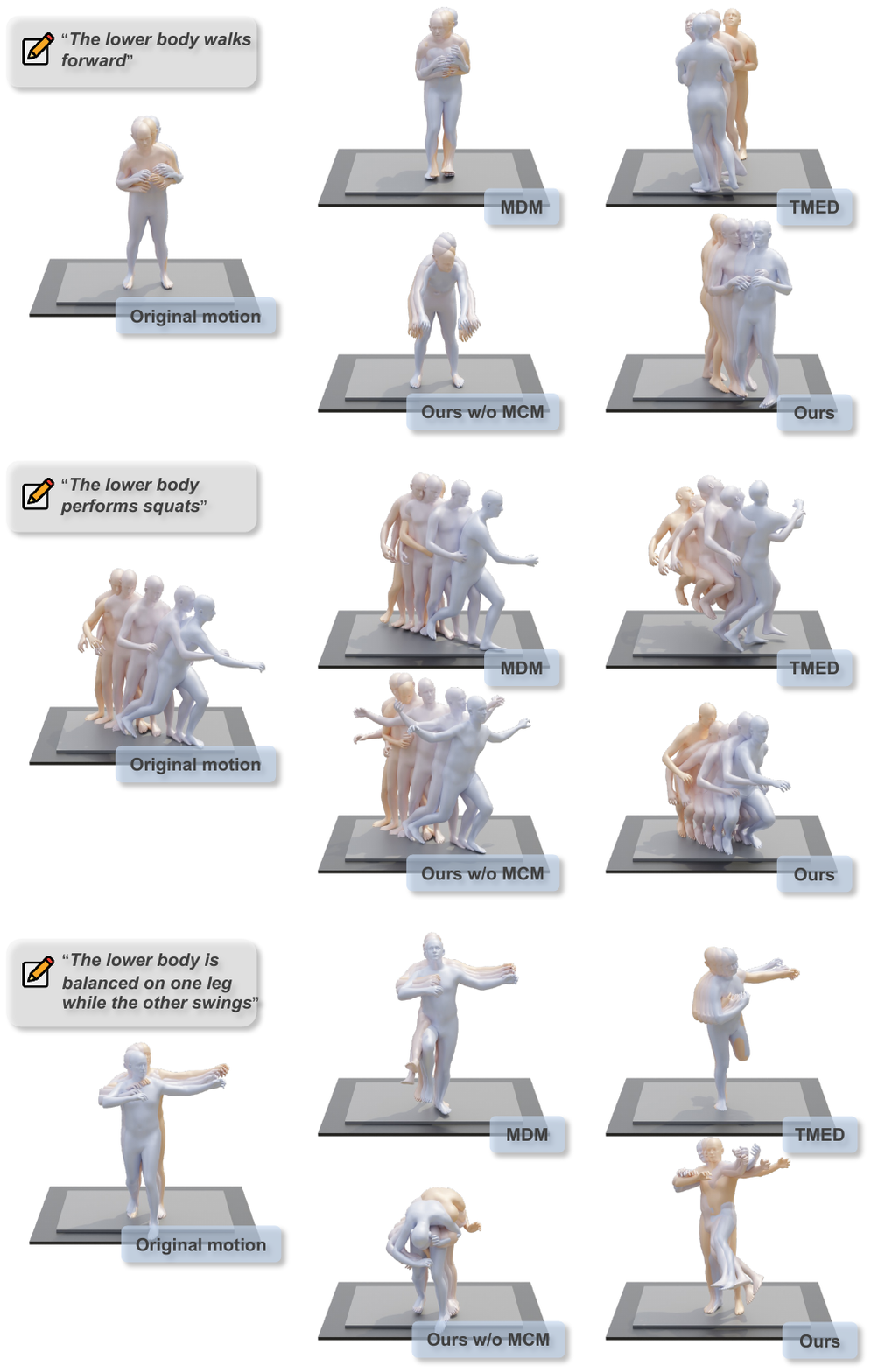}
    \caption{\textbf{Comparison with baselines and ablations on body part replacement.}}
    \label{fig:re_1}
\end{figure*}

\begin{figure*}[ht!]
    \centering
    \includegraphics[width=0.9\textwidth]{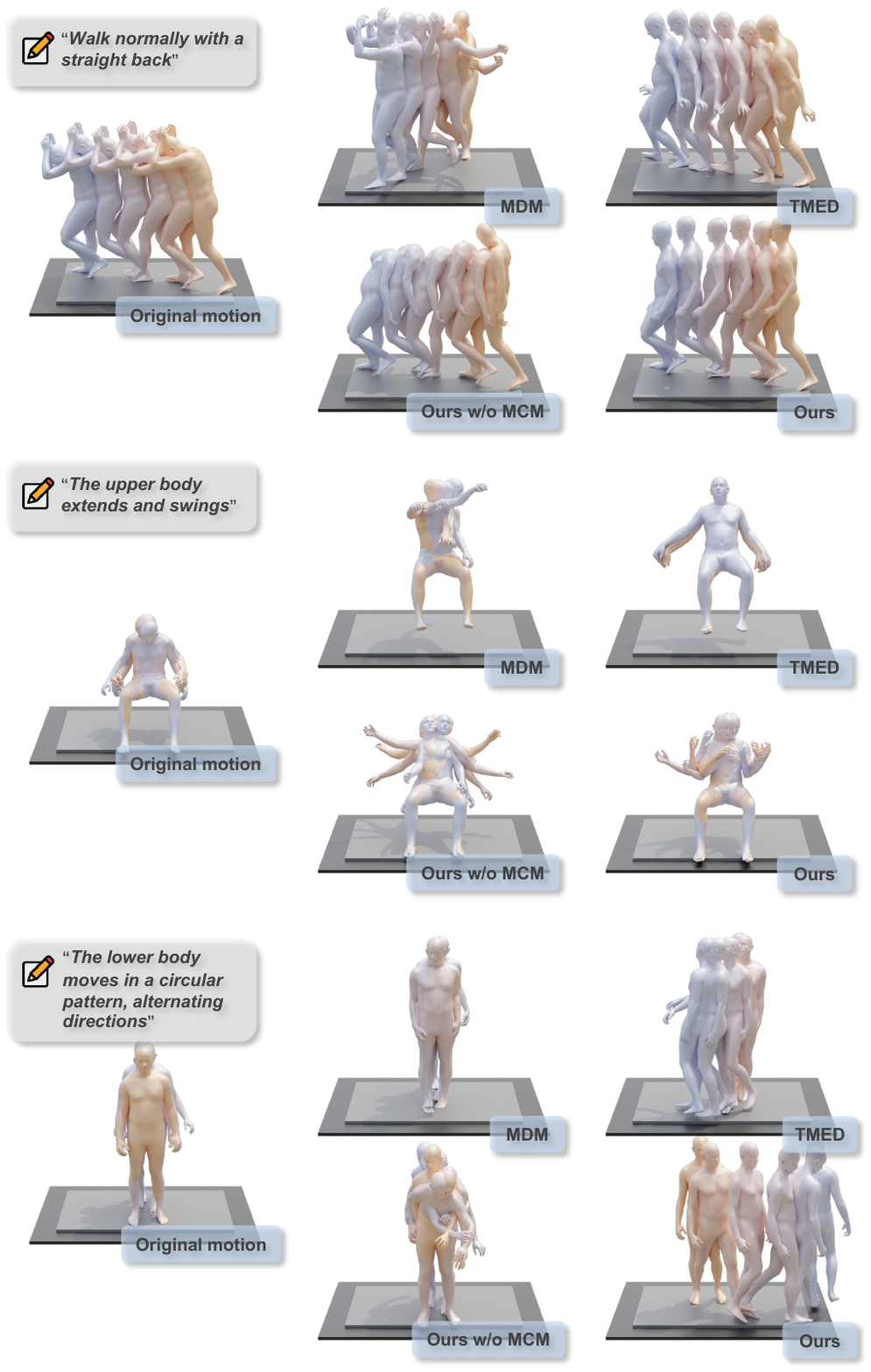}
    \caption{\textbf{Comparison with baselines and ablations on body part replacement.}}
    \label{fig:re_2}
\end{figure*}

\begin{figure*}[ht!]
    \centering
    \includegraphics[width=0.9\textwidth]{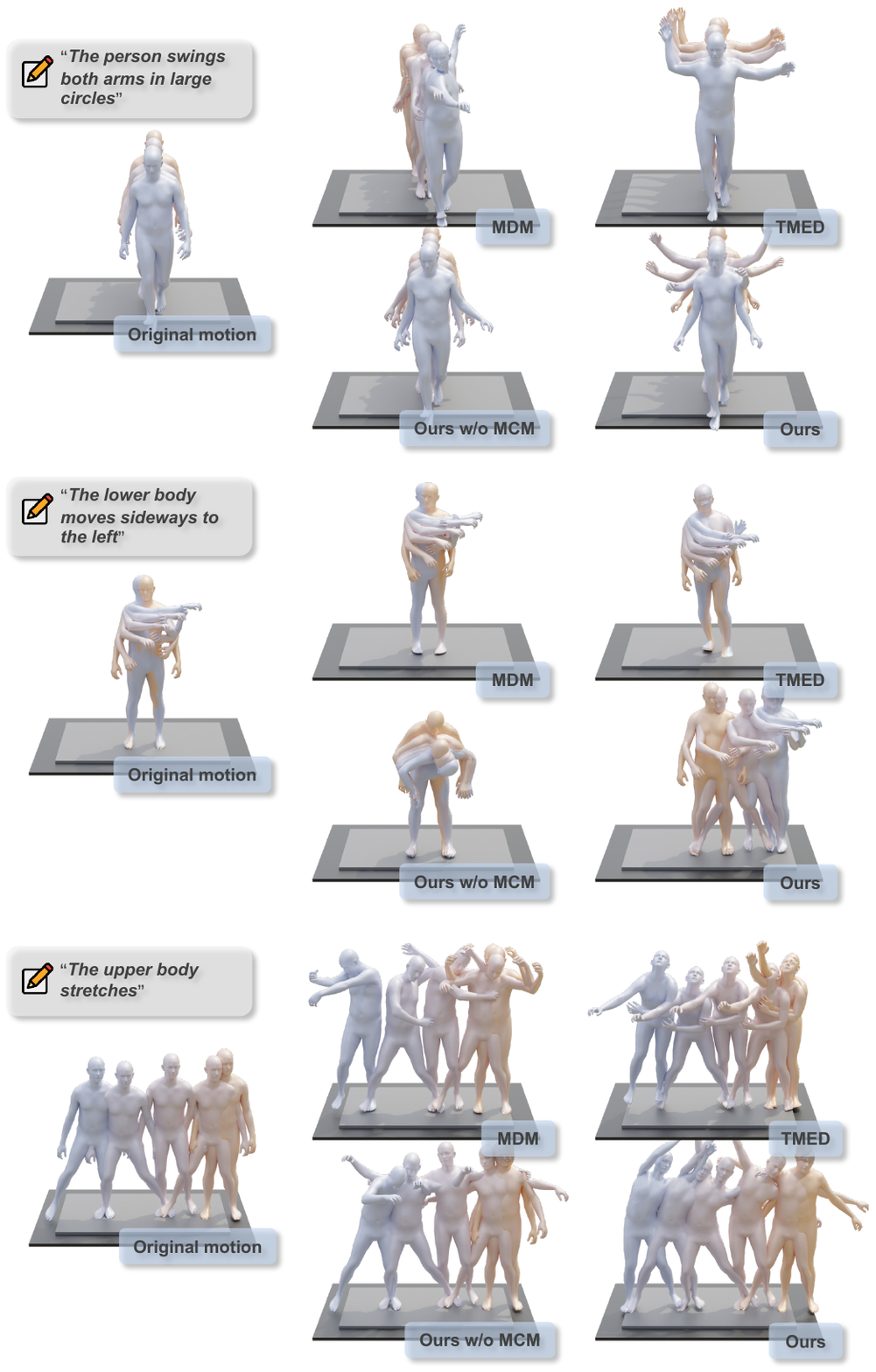}
    \caption{\textbf{Comparison with baselines and ablations on body part replacement.}}
    \label{fig:re_3}
\end{figure*}

\begin{figure*}[ht!]
    \centering
    \includegraphics[width=0.9\textwidth]{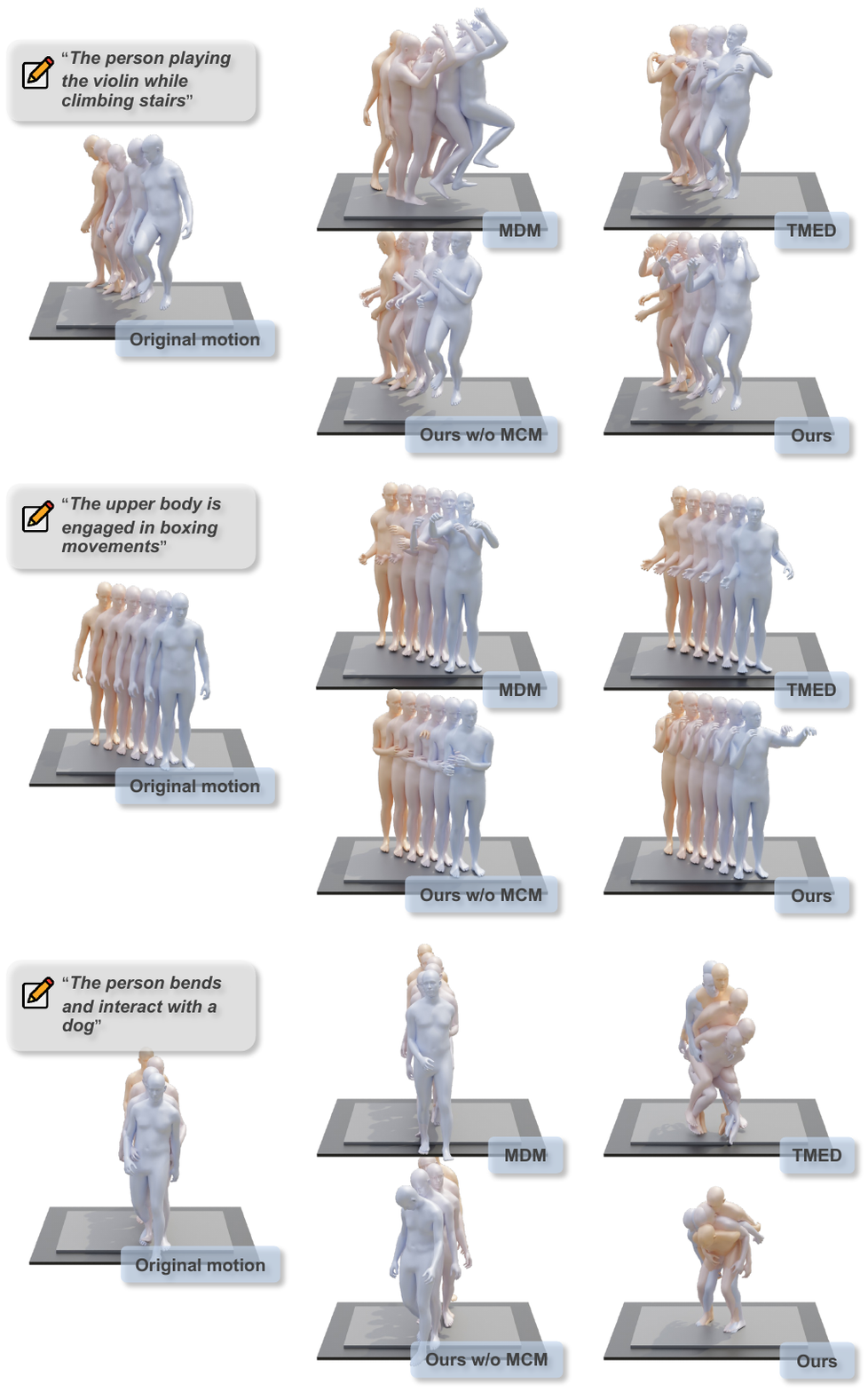}
    \caption{\textbf{Comparison with baselines and ablations on body part replacement.}}
    \label{fig:re_4}
\end{figure*}

We show additional qualitative comparisons for three editing tasks: style transfer (\cref{fig:st_1,fig:st_2,fig:st_3,fig:st_4}), body part replacement (\cref{fig:re_1,fig:re_2,fig:re_3,fig:re_4}), and fine-grained adjustment(\cref{fig:ad_1,fig:ad_2,fig:ad_3}). We highly recommend viewing our \href{https://awfuact.github.io/motionrefit/}{\textit{project website}} for compelling demonstrations across diverse scenarios.

\section{Additional Implementation Details}\label{sec:addtional-implementation-details}

\subsection{Keypoint-Based Motion Representation}\label{sec:keypoint-representation}

Our keypoint-based motion representation uses the first 22 joints from SMPL-X~\cite{pavlakos2019expressive} as primary body joints. The two additional head joints and four finger joints (ring and index fingertips of both hands) correspond to the following SMPL-X indices:
\begin{itemize}
    \item Joint 23: \texttt{left\_eye\_smplhf}.
    \item Joint 24: \texttt{right\_eye\_smplhf}.
    \item Joint 25: \texttt{left\_index1}.
    \item Joint 34: \texttt{left\_ring1}.
    \item Joint 40: \texttt{right\_index1}.
    \item Joint 49: \texttt{right\_ring1}.
\end{itemize}
These additional joints enable natural gaze behavior and head tracking through eye joints, while fingertip joints provide enhanced control over hand poses as end-effectors.

\subsection{Keypoint Canonicalization and Nomalization}\label{sec:keypoint-canonicalization} 

We canonicalize motion segments in a y-up coordinate system to simplify the learning space. For each training segment, we apply a transformation to the entire keypoint sequence that translates the first frame's pelvis to the horizontal origin (x and z) and rotates around the y-axis to align the character's initial forward direction with the positive z-axis. During inference, segments are merged through decanonicalization. Specifically, for segment $i$, we align it with segment $i-1$ by computing the transformation between their connecting frames (first frame of segment $i$ and second-to-last frame of segment $i-1$) using the Kabsch algorithm on the rigid triangle formed by the pelvis and hip joints.

In addition to canonicalization, we normalize each spatial dimension (x, y, and z) of the keypoint data to the standardized range $[-1, 1]$ using channel-specific scaling factors. These factors are determined by the minimum and maximum values of each channel across the dataset. We capture 95\% of the data range to compute these scaling factors with the outliers removed. During inference, we reverse this normalization by applying the inverse scaling factors to the model output.

\subsection{Converting between Motion Representations}\label{sec:joint2smplx}

To convert SMPL-X parameters to keypoint representation, we perform forward kinematics using the official SMPL-X codebase, which transforms sequential pose parameters into 3D joint locations. We set hand and face parameters to zero vectors to focus on core body movements.

Converting keypoint representation to SMPL-X parameters involves a two-stage approach. First, we standardize each frame by translating the 28 keypoints to center the pelvis at the origin. The translated keypoints (84-dimensional input) are processed through a 3-layer MLP (512 hidden units, ReLU activation, layer normalization) to estimate the 66-dimensional SMPL-X body pose parameters, including global orientation. Second, we refine these initial body pose estimates and predict the global translation through optimization. We iteratively compute keypoint locations via SMPL-X forward passes and minimize the mean squared error between the computed and targeted keypoints. Optimization is performed for 120 iterations using the Adam optimizer~\cite{kingma2014adam} with a learning rate of 0.01.

\subsection{Module Details}\label{sec:module-details} 

In our motion diffusion model, noisy motion frames from the canonicalized sequence $\mathcal{M}_t$ are encoded through an MLP encoder, where a single linear layer projects the input from 84 dimensions (28 joints $\times$ 3) to 512 dimensions. The original motion sequence $\mathcal{M}_\text{ori}$ is encoded through a separate MLP encoder with identical architecture. We implement a Transformer encoder~\cite{vaswani2017attention} as the UNet backbone with 6 layers, 16 attention heads, and a dropout rate of 0.1. The encoded vectors from $\mathcal{M}_\text{ori}$ are added frame-wise to the encoded noisy motion to preserve reference motion information. A conditional token combines text condition embedding, progress indicator, and diffusion step embedding for temporal context. The Transformer encoder then processes the entire token sequence, followed by an MLP decoder projecting the output back to 84 dimensions.

For the body part coordinator $D$, we adopt a Transformer encoder with identical architecture to our main model. The transformer's outputs are mean-pooled temporally and processed by an MLP to classify whether the input motion is spatially composed. To ensure robustness during diffusion sampling, we inject random noise into the training keypoint sequences, with magnitude matching the noise levels of the last 20 diffusion steps.

\subsection{Frame Rate}

We downsample motion sequences to 10 FPS during training and inference for computational efficiency. For compatibility with standard evaluation protocols, the generated keypoint sequences are later upsampled to 20 FPS during SMPL-X conversion (\cref{sec:joint2smplx}) to match the original dataset's frame rate.

\subsection{Hyper-Parameters for Guidance} 

During inference, we apply classifier-free guidance~\cite{ho2022classifier} with weight $w=3$ to enhance conditional signals through linear extrapolation. For the body part coordinator, we set $\lambda$ to 1.0 and apply classifier guidance during the final 20 steps of the auto-regressive sampling process.

\section{Additional Experiment Details and Results}\label{sec:experiments-supp}

\subsection{Training Details} 

In our experimental framework, all models undergo training for 1,500 epochs using the DDPM scheduler~\cite{ho2020denoising}, with varying numbers of diffusion steps across different methods: our approach employs 100 steps, TMED~\cite{athanasiou2024motionfix} uses 300, and MDM-BP~\cite{tevet2022mdm} requires 1,000, following their respective recommended configurations. We employ the AdamW optimizer~\cite{loshchilov2017decoupled} with a learning rate of 1e-4 and a weight decay of 0.01. The learning rate follows a linear decay schedule. During training, we use a batch size of 1024 sequences, with each sequence containing $W$ frames. The training process is conducted on a setup of 4 NVIDIA RTX 3090 GPUs, with the entire training cycle completed within 36 hours. The model checkpoints are saved every 50 epochs, and we select the best model based on validation performance.

\subsection{Adaption of Baselines}\label{sec:baseline-adaption}

For baseline comparisons, we adapt MDM~\cite{tevet2022mdm} with inpainting-based motion editing, where specific body parts are modified according to the provided masks. We enhance the baseline by supplying explicit masking information and initializing diffusion from the original motion sequence. We introduce an important modification to the standard MDM approach: while most of the diffusion process maintains strict masking constraints, we release these constraints during the final 20 diffusion steps, allowing the model to adjust the entire body. This modification enables natural whole-body adaptations that may be necessary for coherent motion synthesis. For TMED~\cite{athanasiou2024motionfix}, we maintain strict adherence to the original implementation, utilizing the exact configurations and parameters as specified in the authors' codebase.

\begin{table*}[ht!]
    \centering
    \small
    \setlength{\tabcolsep}{6.8pt}
    \caption{\textbf{Quantitative comparison on fine-grained adjustment task.} For each metric, we repeat the evaluation 10 times. Arrows ($\rightarrow$) indicate metrics where values closer to real data are better. \textbf{Bold} denotes best performance.}
    \label{tab:results-adjustment}
    \begin{tabular}{lccccccccccc}
        \toprule
        \multirow{2}{*}{Method} & \multirow{2}{*}{FID$\downarrow$} & \multirow{2}{*}{Diversity$\rightarrow$} & \multirow{2}{*}{FS$\uparrow$} & \multicolumn{4}{c}{Edited-to-Source Retrieval} & \multicolumn{4}{c}{Edited-to-Target Retrieval} \\
        \cmidrule(r){5-8} \cmidrule(l){9-12} & & & & R@1$\uparrow$ & R@2$\uparrow$ & R@3$\uparrow$ & AvgR$\downarrow$ & R@1$\uparrow$ & R@2$\uparrow$ & R@3$\uparrow$ & AvgR$\downarrow$ \\
        \midrule
        Real Data & 0.02 & 30.57 & 0.97 & 39.54& 54.65& 61.16& 5.53& 100.0& 100.0& 100.0& 1.00\\
        MDM-BP~\cite{tevet2022mdm} & 0.62 & 32.70 & 0.92 & 28.12& 34.38& 38.02& 10.41 &16.45& 24.52& 30.21& 11.60 \\
        TMED~\cite{athanasiou2024motionfix} & 0.33 & 31.13& 0.94 & 60.16& 72.66& 82.03& 2.66& 29.69& 44.01& 52.08& 6.97 \\
        TMED w/ MCM & 0.33 & 31.42 & 0.94 & 62.8& 74.78& 87.0& 2.61& 32.22& 45.03& 54.83& 6.56\\
        Ours w/o MCM & 0.34& \textbf{31.08} & \textbf{0.95} & 81.77& 92.45& 93.49& 1.48& 34.11& 48.70& 57.03& 5.77 \\
        Ours full & \textbf{0.29}& 31.29 & \textbf{0.95} & \textbf{85.16}& \textbf{92.97}& \textbf{95.31}& \textbf{1.38}& \textbf{42.45}& \textbf{56.25}&  \textbf{62.76} &\textbf{5.12}\\
        \bottomrule
    \end{tabular}
\end{table*}

\begin{table*}[ht!]
    \centering
    \small
    \setlength{\tabcolsep}{8pt}
    \caption{\textbf{Ablation analysis for fine-grained adjustment.} Results show means across 10 evaluation runs, with \textbf{bold} indicating best result.}
    \label{tab:motion-adjustment}
    \begin{tabular}{lccccccccc}
        \toprule
        \multirow{2}{*}{Method} & \multirow{2}{*}{FID$\downarrow$} & \multicolumn{4}{c}{Edited-to-Source Retrieval} & \multicolumn{4}{c}{Edited-to-Target Retrieval} \\
        \cmidrule(r){3-6} \cmidrule(l){7-10} & & R@1$\uparrow$ & R@2$\uparrow$ & R@3$\uparrow$ & AvgR$\downarrow$ & R@1$\uparrow$ & R@2$\uparrow$ & R@3$\uparrow$ & AvgR$\downarrow$ \\
        \midrule
        1\% MCM & 0.34 & 81.77 & 92.45 & 93.49 & 1.48 & 34.11 & 48.70 & 57.03 & 5.77 \\
        5\% MCM & 0.37 & \textbf{86.72} & \textbf{95.57} & \textbf{97.14} & \textbf{1.30} & 34.17 & 50.00 & 57.81 & 5.65 \\
        10\% MCM  & 0.31 & 82.81 & 92.71 & 95.31 & 1.42 & 37.24 & 51.30 & 59.11 & 5.32 \\
        20\% MCM  & 0.29 & 85.68 & 91.93 & 94.27 & 1.45 & 39.06 & 52.08 & 60.68 & 5.36 \\
        12\% data & 0.32 & 81.51 & 91.67 & 94.53 & 1.56 & 40.10 & 58.07 & \textbf{67.71} & 4.74 \\
        24\% data & 0.31 & 82.03 & 92.19 & 95.83 & 1.42 & 41.93 & \textbf{59.11} & 67.45 & \textbf{4.71} \\
        60\% data & 0.30 & 84.90 & 92.45 & 96.09 & 1.38 & 41.67 & 55.47 & 63.54 & 5.02 \\
        Ours full & \textbf{0.29}& 85.16 & 92.97 & 95.31 & 1.38 & \textbf{42.45} & 56.25 & 62.76 & 5.12 \\
        \bottomrule
    \end{tabular}
\end{table*}

\begin{table*}[ht!]
    \centering
    \small
    \setlength{\tabcolsep}{7pt}
    \caption{\textbf{Quantitative comparison with TMED~\cite{athanasiou2024motionfix} on MotionFix using the full dataset~\cite{athanasiou2024motionfix}.} Results show means across 10 evaluation runs, with \textbf{bold} indicating best result.}
    \label{tab:tmed-full}
    \begin{tabular}{lcccccccccc}
        \toprule
        \multirow{2}{*}{Method} & \multirow{2}{*}{FID$\downarrow$} & \multirow{2}{*}{FS$\uparrow$} & \multicolumn{4}{c}{Edited-to-Source Retrieval} & \multicolumn{4}{c}{Edited-to-Target Retrieval} \\
        \cmidrule(r){4-7} \cmidrule(l){8-11} & & & R@1$\uparrow$ & R@2$\uparrow$ & R@3$\uparrow$ & AvgR$\downarrow$ & R@1$\uparrow$ & R@2$\uparrow$ & R@3$\uparrow$ & AvgR$\downarrow$ \\
        \midrule
        Real Data                           & 0.010  & 0.98  & 20.83 & 33.66 & 40.47 & 33.13 & 64.36 & 88.75 & 95.56 & 1.74 \\
        MDM-BP~\cite{tevet2022mdm}          & 0.145  & 0.90  & 30.21 & 36.82 & 40.47 & 106.05 & 8.69 & 14.71 & 18.36 & 180.99  \\
        TMED~\cite{athanasiou2024motionfix} & 0.129  & 0.92  & 22.41 & 34.45 & 40.57 & 31.42 & \textbf{14.51} & 21.72 & 28.73 & 56.63 \\
        Ours                                & \textbf{0.120}  & \textbf{0.96}  & \textbf{43.77} & \textbf{56.72} & \textbf{64.13} & \textbf{24.09}  & 14.13 & \textbf{23.52} & \textbf{30.53} &  \textbf{54.06}  \\
        \bottomrule
    \end{tabular}
\end{table*}

\subsection{Dual Interpretation of the E2S Score}\label{sec:e2s-score}

We argue that the interpretation of Edited-to-Source Retrieval (E2S) scores should be task-dependent.

For fine-grained adjustments (\eg, modifying arm raise height), higher E2S scores are desirable as they indicate preserved motion characteristics with successful subtle modifications. Similarly, for MotionFix dataset~\cite{athanasiou2024motionfix} tasks which involve minor adjustments like refining limb positions and trajectories, high E2S scores demonstrate proper maintenance of source motion semantics.

However, for substantial editing tasks like body part replacement or style transfer, the E2S scores should align with the reference dataset's distribution rather than maximizing similarity to the source. In these cases, lower E2S scores may actually indicate successful editing, as the motion should significantly deviate from the source to reflect the intended modifications. The accuracy of these major changes should instead be evaluated through the Edited-to-Target Retrieval score, which measures alignment with the target characteristics.

\subsection{Ablation Results of Classifier Guidance}\label{sec:guidance}

\begin{figure}[h!]
    \centering
    \begin{subfigure}{\linewidth}
        \centering
        \includegraphics[width=\linewidth]{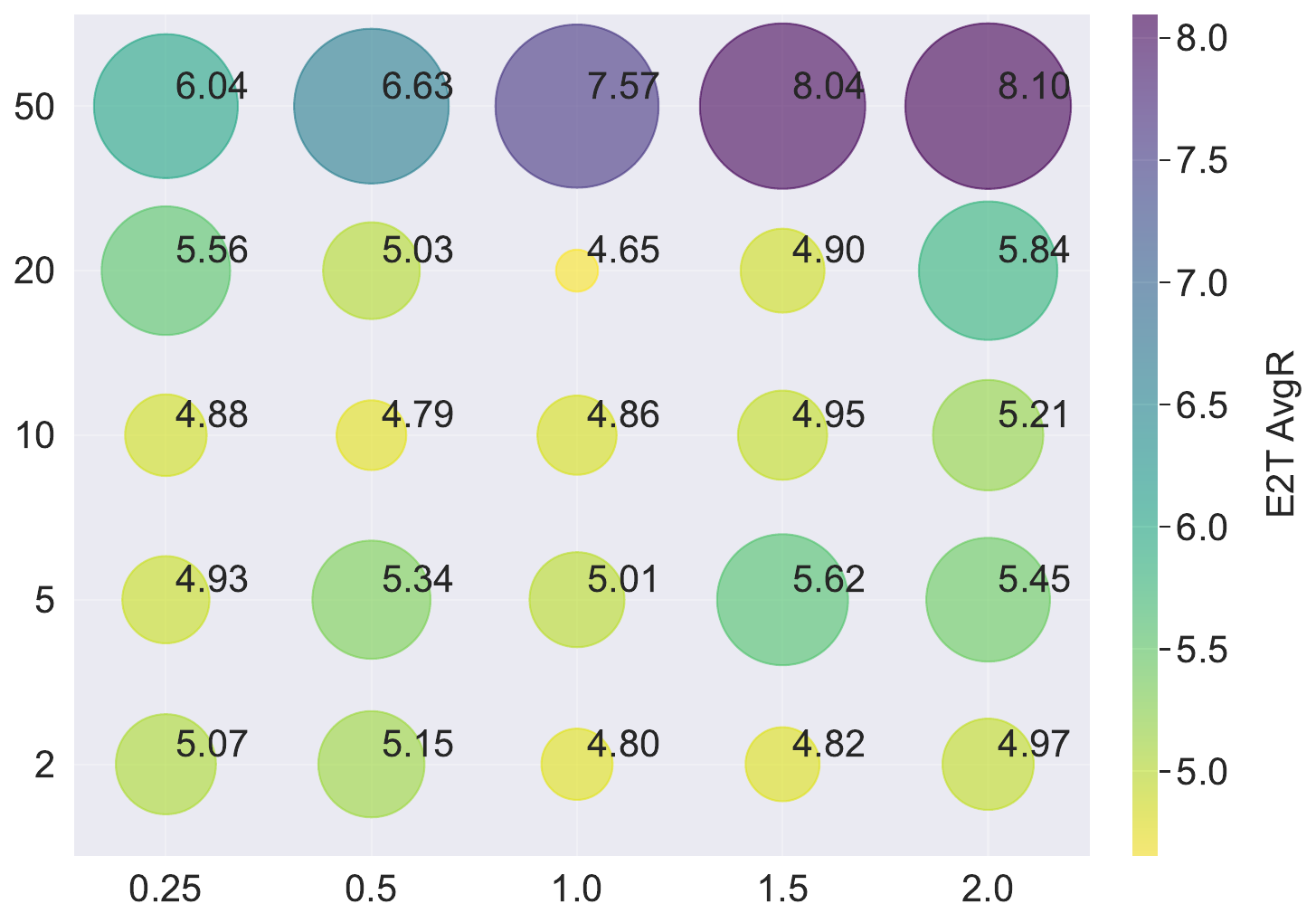}
    \end{subfigure}
    \\
    \begin{subfigure}{\linewidth}
        \centering
        \includegraphics[width=\linewidth]{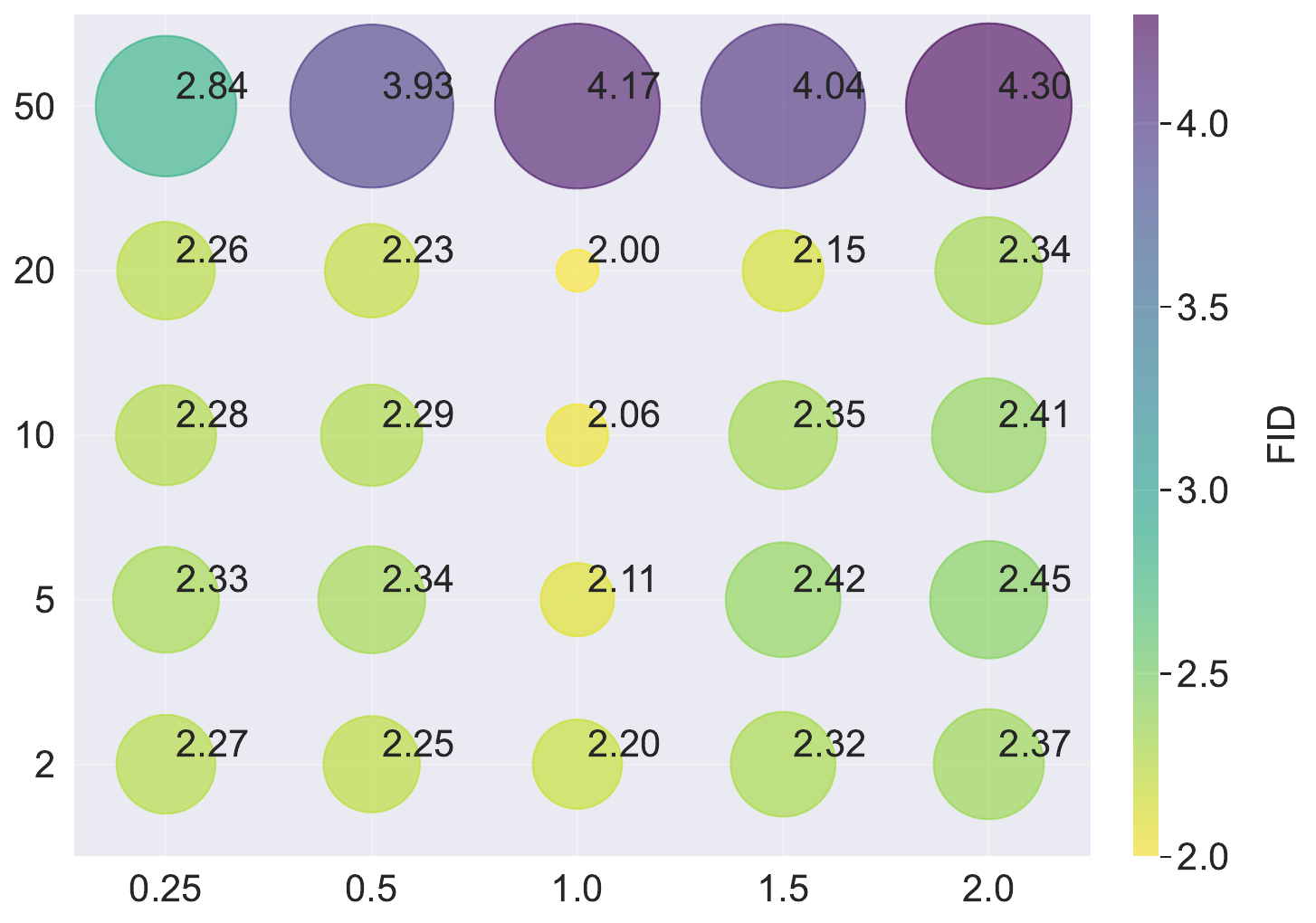}
    \end{subfigure}
    \caption{\textbf{Ablation results on classifier guidance.} We illustrate the E2T AvgR (upper) and FID (lower) performance of \model for the body part replacement task. The x-axis represents guidance strength, whereas the y-axis depicts guidance steps count.}
    \label{fig:ablation-guidance}
\end{figure}

In \cref{fig:ablation-guidance}, we evaluate how body part coordinator performs across different hyper-parameters. The x-axis shows guidance strength $\lambda$, while the y-axis indicates the number of steps where classifier guidance is applied. We report both E2T AvgR (upper) and FID (lower) for the body part replacement task. Setting $\lambda=1.0$ and applying 20 guidance steps produces optimal results.

\subsection{Results of Fine-Grained Adjustment}\label{sec:results-of-adjustment}

Quantitative results in \cref{tab:results-adjustment} demonstrate that our full method achieves superior performance across most metrics for the fine-grained adjustment task. The retrieval metrics reveal that the motion characteristics have been maintained, with successful fine-grained adjustments.

\cref{tab:motion-adjustment} presents quantitative comparisons between our method and ablation variants on the fine-grained adjustment task. While increasing the MotionCutMix Ratio generally enhances results, we find that a lower ratio of 5\% actually achieves optimal performance, outperforming higher ratios including 100\%. This phenomenon can be attributed to the inherent consistency of editing patterns across fine-grained motion adjustments. Additionally, our experiments show that varying the size of the annotated dataset produces only marginal differences in performance metrics. This finding suggests that our method achieves effective generalization even with a smaller annotated dataset, likely because our large-scale training set already encompasses a comprehensive range of fine-grained adjustment scenarios.

\begin{figure*}[ht!]
    \centering
    \includegraphics[width=0.9\textwidth]{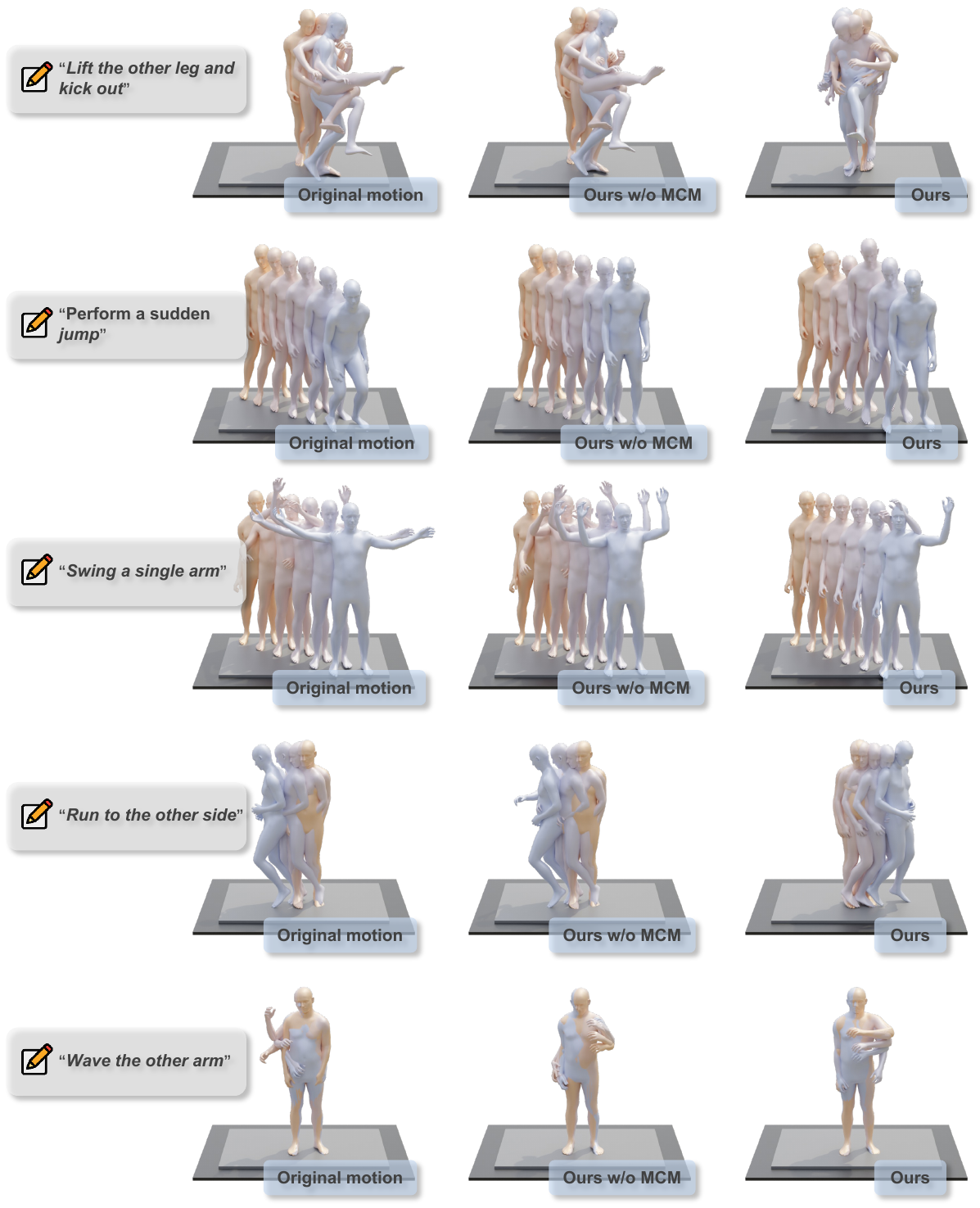}
    \caption{\textbf{Comparison with ablations on fine-grained adjustment.}}
    \label{fig:ad_1}
\end{figure*}

\begin{figure*}[ht!]
    \centering
    \includegraphics[width=0.9\textwidth]{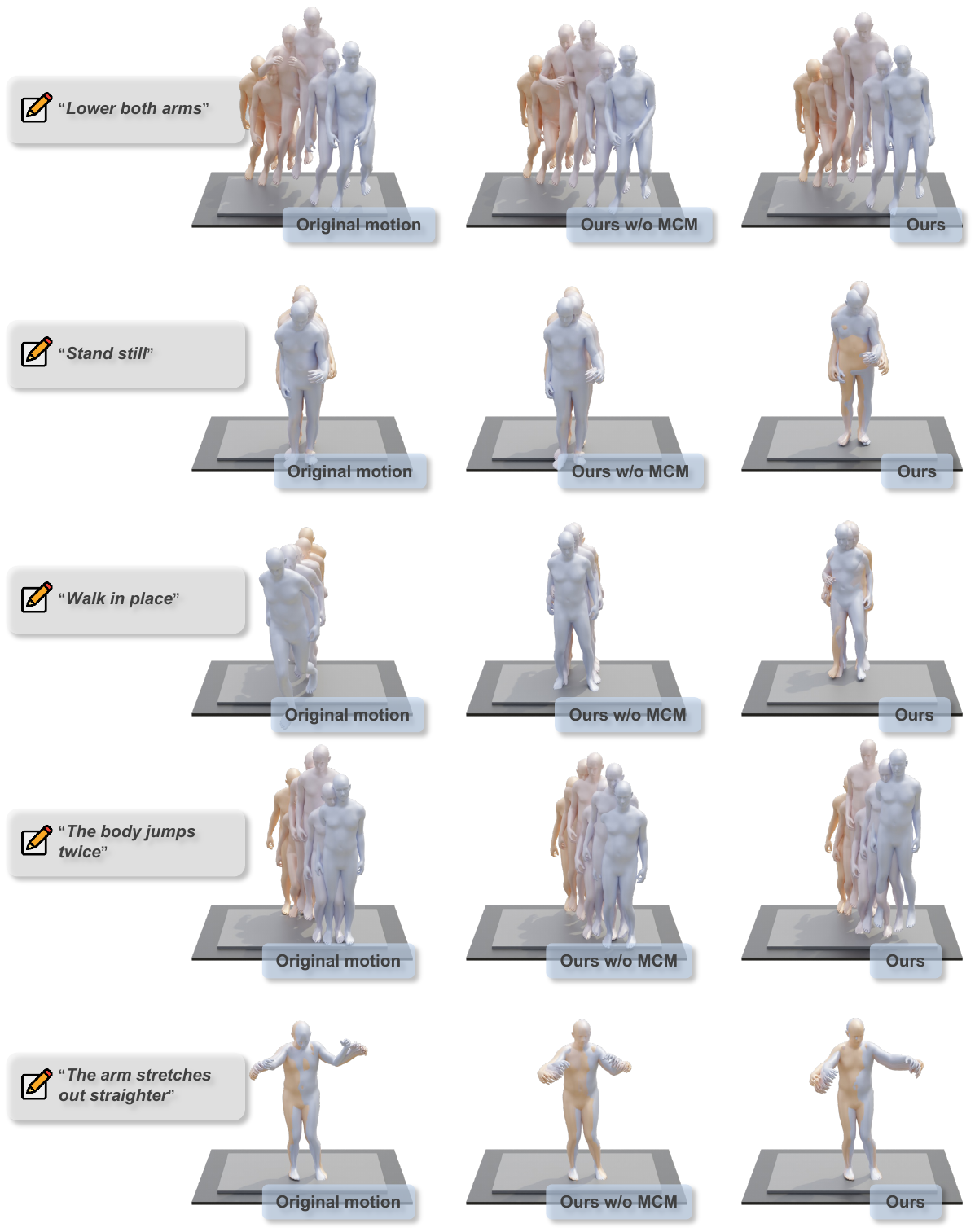}
    \caption{\textbf{Comparison with ablations on fine-grained adjustment.}}
    \label{fig:ad_2}
\end{figure*}

\begin{figure*}[ht!]
    \centering
    \includegraphics[width=0.9\textwidth]{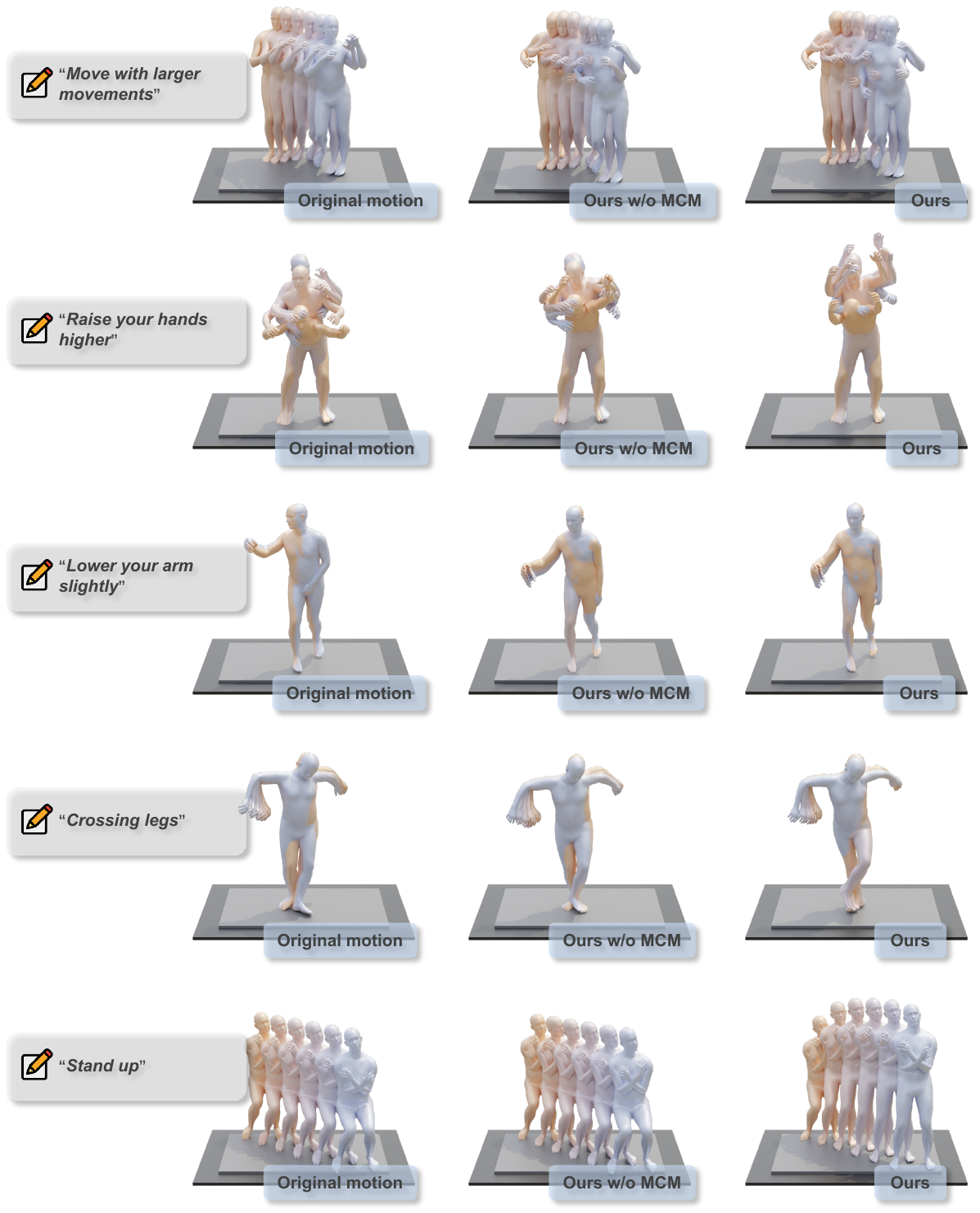}
    \caption{\textbf{Comparison with ablations on fine-grained adjustment.}}
    \label{fig:ad_3}
\end{figure*}

\cref{fig:ad_1,fig:ad_2,fig:ad_3} showcase visual comparisons between our method and ablations across diverse editing instructions, demonstrating our full method's superiority in producing precise and natural motion edits.

\subsection{Results on the MotionFix Dataset}\label{sec:detail-and-results-on-motionfix}

\paragraph{Evaluation Settings} 

For TMED~\cite{athanasiou2024motionfix} compatibility, we use a 22-keypoint representation aligned with the SMPL model~\cite{loper2015smpl}, instead of the 28-keypoint SMPL-X format used in our main method.  The conversion process between keypoint representation and SMPL parameters remains similar to the one described in \cref{sec:joint2smplx}.

For our auto-regressive framework, we preprocess the MotionFix dataset by segmenting continuous motions into clips and applying canonicalization. For retrieval-based metrics evaluation, we use the original TMR checkpoint~\cite{petrovich2023tmr} to ensure consistent comparison with previously reported results.

\paragraph{Comparison on the Entire Test Set}

\cref{tab:tmed-full} shows full-scale evaluation results on the MotionFix benchmark comparing our method against TMED and MDM baselines. Consistent with the batch-wise evaluation, our method demonstrates superior performance in both E2T scores for editing accuracy and E2S scores for motion preservation. Most notably, we achieve substantially higher foot contact scores, indicating significantly improved physical plausibility and overall motion quality.

For detailed qualitative comparisons and motion visualizations that further illustrate these improvements, we direct readers to \cref{sec:additional-results}.

\begin{table}[t!]
    \centering
    \small
    \caption{\textbf{Breakdown of inference time on a single RTX 3090 GPU.} Our optimal setting achieves real-time inference speed.}
    \label{tab:inference-speed}
    \resizebox{\linewidth}{!}{%
        \begin{tabular}{lccccc}
            \toprule
            \begin{tabular}[c]{@{}c@{}}Window \\ size\end{tabular} & \begin{tabular}[c]{@{}c@{}}Diffusion \\ sampling\end{tabular} & \begin{tabular}[c]{@{}c@{}}Body part \\ coordinator\end{tabular} & \begin{tabular}[c]{@{}c@{}}SMPL-X \\ optimization\end{tabular} & \begin{tabular}[c]{@{}c@{}}Total \\ (seconds)\end{tabular} & FPS \\
            \midrule
            2-frame  & 0.142& 0.014 & 0.067 & 0.223& 8.97\\
            8-frame  & 0.355& 0.036 & 0.106 & 0.497& 16.10\\
            16-frame & 0.474& 0.046 & 0.126 & 0.646& 24.76\\
            \bottomrule
        \end{tabular}
    }
\end{table}

\subsection{Real-Time Inference}

In \cref{tab:inference-speed}, we provide a breakdown of inference time on a single RTX 3090 GPU. Despite the auto-regressive nature, inference with a 16-frame window size (our optimal setting) achieves real-time speed. Furthermore, the motion coordinator is applied only during the final few diffusion steps, adding minimal overhead to the overall computation.

\section{Additional Details on the STANCE Dataset} \label{sec:addtional-dataset-details}

\subsection{Body Part Replacement}

Our body part replacement subset extends HumanML3D~\cite{guo2022humanml3d} through a two-phase annotation process capturing both body part participation and detailed motion descriptions.

\paragraph{Mask Annotation} 

\begin{figure}[t!]
    \centering
    \includegraphics[width=0.43\textwidth]{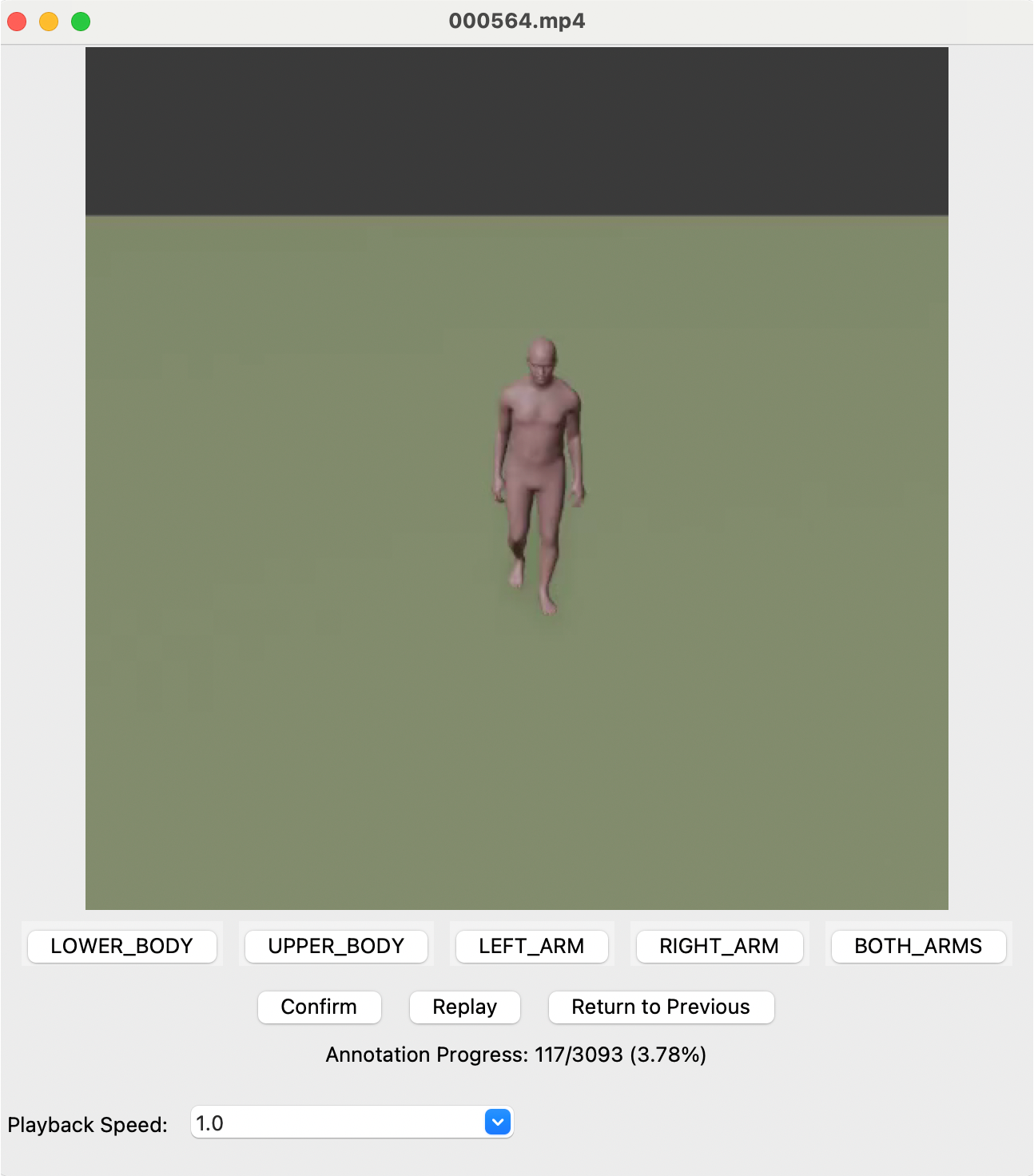}
    \caption{\textbf{Screenshot of our annotation software.}}
    \label{fig:software}
\end{figure}

The first phase focuses on mask annotation, where we developed specialized visualization software to streamline the annotation process. As shown in \cref{fig:software}, this tool renders HumanML3D motion sequences in 3D and offers annotators a selection of predefined body part masks and their combinations. Annotators can play, pause, and scrub through the animation while making their selections based on direct visualization of the motions. For each sequence, annotators identify which body parts are actively participating in meaningful movements, as opposed to parts that remain relatively static or perform only supporting motions. This visual-based annotation approach distinguishes our dataset from previous works that rely solely on language model interpretation of text descriptions to determine body part involvement~\cite{athanasiou2023sinc}. We employed five trained annotators who processed sequences from HumanML3D, resulting in multiple mask annotations per sequence.

\paragraph{Detailed Annotation} 

The second phase involves creating detailed descriptions for the movements of designated body parts. We initialize this process using GPT-4 to obtain the original HumanML3D motion descriptions and specific instructions to focus on particular body parts while excluding others. For example, given a motion described as ``a person walks forward while waving their arms,'' and focusing on the arms, the LLM might generate ``waves arms enthusiastically from side to side.'' These initial descriptions then undergo careful refinement by human annotators who enhance their accuracy, naturalness, and linguistic diversity. This combined approach leverages both automated assistance and human expertise to create approximately 13,000 rich, precise annotations of body part movements. Each motion sequence receives 2-4 different body part-specific descriptions, creating a diverse set of potential editing targets.

\begin{figure*}[b!]
    \centering
    \includegraphics[width=\textwidth]{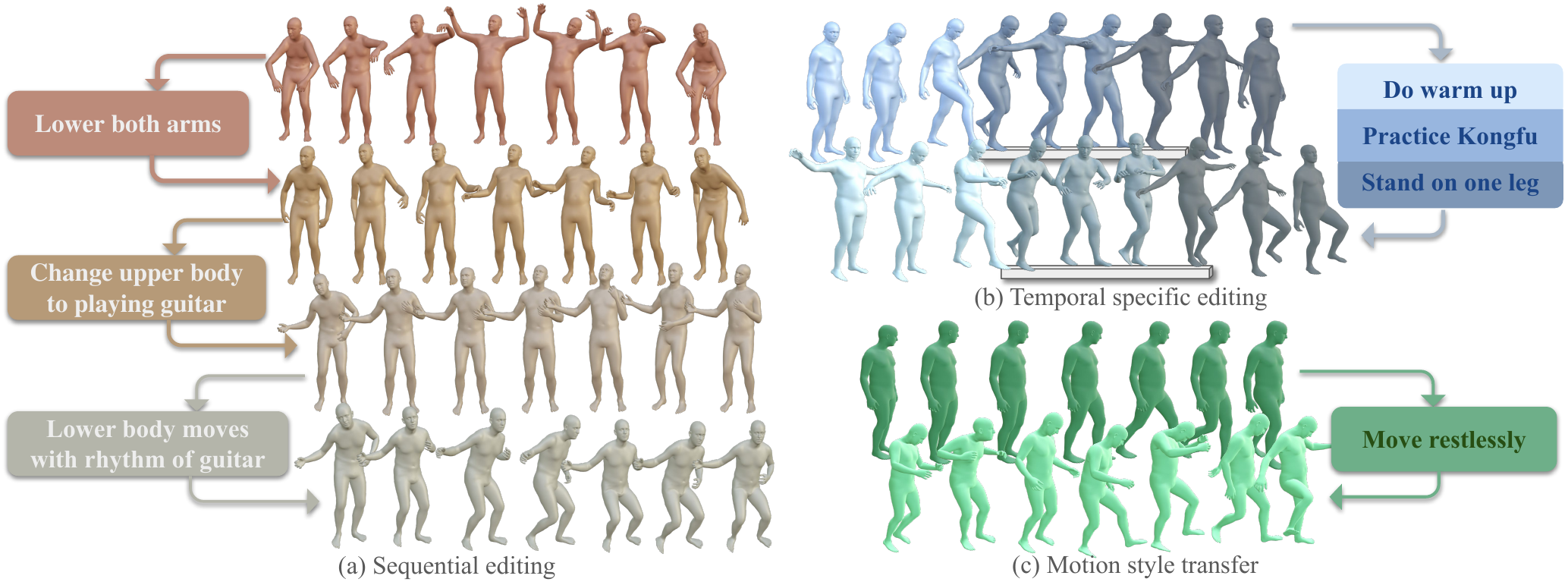}
    \caption{\textbf{Compositional applications performed by our method.}}
    \label{fig:application}
\end{figure*}

\subsection{Motion Style Transfer}

We construct a motion style transfer dataset by professionally recreating sequences from HumanML3D~\cite{guo2022humanml3d} using the high-fidelity Vicon motion capture system. In our capture sessions, we enlisted trained performers who were instructed to replay selected HumanML3D sequences while incorporating specific style variations. They first familiarized themselves with the original motions through video playback and practice sessions, then executed each motion multiple times with different stylistic interpretations. The capture setup consisted of 12 Vicon cameras operating at 30 fps, positioned strategically around a $6\times6$ meter capture volume. Performers wore a standard 53-marker FrontWaist set for full-body tracking, ensuring accurate capture of subtle stylistic nuances. 

We focused on distinct style categories: proud, old, playful, depressed, and angry, with each performer interpreting these styles based on provided style guidelines. From 180 base motions selected from HumanML3D, we captured each motion in all five styles, resulting in a dataset of 900 high-quality motion sequences after post-processing and cleanup. Each sequence is paired with its original HumanML3D counterpart and annotated with detailed descriptions of the stylistic differences, creating style transfer triplets suitable for training and evaluation.

\subsection{Fine-Grained Adjustment}

We introduce a novel text-to-motion generation approach for obtaining semantically consistent motion pairs. We curate 5,000 base instructions spanning common human actions (walking, running, dancing, sports activities). For each instruction, we generate the initial motion using MLD's standard sampling process~\cite{chen2023executing}. To create variants, we additionally apply Gaussian noise ($\sigma$ = 0.1) to the latent space, creating 16 slightly different but semantically consistent variations for each base motion. These variants maintain the core action while exhibiting subtle differences in execution style, speed, or range of movement.

The variants are then paired one-to-one, creating 8 pairs per instruction. Trained annotators carefully examine each pair and describe the specific modifications needed to transform the original motion into the edited motion. The annotations focus on precise, actionable descriptions such as ``bend the knees more deeply,'' ``perform the arm swing with greater force,'' or ``slow down the spinning movement slightly.'' To ensure dataset quality and clarity, we implement a rigorous filtering process where triplets with unnatural motions (\eg, physically implausible movements) or unclear editing descriptions are discarded. Additionally, we maintain a balanced distribution across different motion categories and editing types to prevent dataset bias.

This systematic approach results in a large-scale dataset of 16,000 annotated triplets, each consisting of an original motion, an edited motion, and a clear instruction for the required modification. The dataset covers a wide range of fine-grained adjustments, including changes in motion amplitude, speed, force, and spatial positioning of body parts.

\section{Compositional Applications}\label{sec:applications}

As shown in \cref{fig:application}, our method enables both interactive editing and complex compositional motion generation, advancing beyond simple motion modifications. This capability distinguishes our approach from prior works that address only specific editing scenarios or isolated modifications.

\subsection{Time-Variant Motion Editing} 

We enable time-variant motion editing through different text instructions. Users can independently modify distinct motion segments by applying different instructions to specific frame ranges. For instance, users can specify ``raise right hand higher'' for the first 25 frames, followed by ``lower the right hand'' for subsequent frames. This fine-grained control is implemented by iteratively calling the auto-regressive model with the first instruction until frame 25, then continuing with the second instruction from frame 25 onward.

\subsection{Interactive Motion Modification} 

Our model supports interactive motion modification by using previously edited motions as input for subsequent processes. Users can build upon earlier edits by feeding the modified motion back into the model with new instructions. For example, after raising an arm, users can further adjust its position by applying additional modifications to the edited motion. This sequential editing process enables progressive refinement until the desired motion is achieved.

\subsection{Compositional Motion Generation} 

Our model enables compositional motion generation through time-variant motion editing and interactive motion modifications. Starting with a base motion, users can layer multiple actions by applying sequential edits. For instance, to create a motion of simultaneously drinking water and reading, users first modify a standing pose with ``drink water'' followed by ``reading the book using the other hand'' applied to the resulting motion.

\section{Limitations and Future work}\label{sec:limitations}

While our method demonstrates strong performance across various editing tasks, it does have several notable limitations that warrant discussion. (i) Our approach shows reduced effectiveness when handling complex temporal dependencies in motion sequences, such as sequential actions (\eg, a number of crouch-stand cycles). (ii) Our model struggles with instructions that require comprehension of spatial relationships (\eg, return to the starting point after forward movement). (iii) While the model performs well on editing patterns similar to those in the training data, its behavior with novel or significantly different editing instructions remains unexplored. 

Future work could focus on: (i) Enhancing the model's spatial-temporal understanding to better handle more complex motion sequences and editing instructions (\eg, adopting motion representations from works that separately consider body parts). (ii) Incorporating physics-based constraints to ensure physical plausibility in extreme editing cases.

\end{document}